\documentclass[compsoc,conference,a4paper,10pt,times]{IEEEtran}
\IEEEoverridecommandlockouts

\usepackage[dvipsnames,table]{xcolor}
\usepackage{colortbl}
\usepackage{xspace}
\usepackage{subcaption}
\usepackage{amsmath}
\usepackage{amssymb}

\usepackage{listings}

\usepackage{pifont}
\usepackage{amsthm}
\theoremstyle{definition}

\usepackage{multirow}
\usepackage{multicol}

\usepackage[most]{tcolorbox}

\usepackage{hhline}

\usepackage{graphicx}
\usepackage{tikz}
\usepackage[inline,shortlabels]{enumitem}
\usepackage{textcomp}
\usepackage{booktabs}
\usepackage{microtype}

\PassOptionsToPackage{hyphens}{url}
\usepackage{url}
\usepackage{hyperref}
\definecolor{darkblue}{rgb}{0, 0, 0.5}
\hypersetup{colorlinks=true, citecolor=darkblue, linkcolor=darkblue,
            urlcolor=darkblue}
\usepackage{cleveref}

\usepackage{float}

\usepackage{makecell}
\usepackage{array}
\usepackage{tabularx}

\newcommand{\system}{CortexCTI\xspace}
\newcommand{\code}[1]{\texttt{#1}}

\newcommand{\stitle}[1]{\vspace{1mm}\noindent\textbf{#1}}
\newcommand{\rev}[1]{#1} 
\definecolor{CadetBlue}{RGB}{95,158,160}

\newcommand{\best}[1]{\textbf{#1}}
\newcommand{\second}[1]{\underline{#1}}

\renewcommand{\arraystretch}{1.1}

\newcommand{\papertablestyle}{%
  \centering
  \footnotesize
  \renewcommand{\arraystretch}{1.08}%
  \setlength{\tabcolsep}{3.25pt}%
  \setlength{\aboverulesep}{1pt}%
  \setlength{\belowrulesep}{1.5pt}%
  \captionsetup{position=top,skip=2pt,justification=raggedright,singlelinecheck=false}%
}
\newcolumntype{L}{>{\raggedright\arraybackslash}X}
\newcolumntype{C}{>{\centering\arraybackslash}X}

\definecolor{darkgreen}{RGB}{0, 100, 0}
\definecolor{experimentblue}{RGB}{127, 159, 186}
\definecolor{experimentgreen}{RGB}{127, 186, 130}
\definecolor{experimentred}{RGB}{186, 138, 127}
\definecolor{darkorange}{RGB}{240,131,15}

\newtcolorbox[auto counter]{findings}[1]{
  breakable, boxrule=0pt, boxsep=3pt,left=0pt,right=0pt,top=0pt,bottom=0pt,
  colbacktitle=experimentgreen, title= \thesubsection{} #1: Findings,
}

\newtcolorbox[auto counter]{defbox}[2]{
  breakable, boxrule=0pt, boxsep=3pt,left=0pt,right=0pt,top=0pt,bottom=0pt,
  colbacktitle=CadetBlue!45, coltitle=black, fonttitle=\sffamily\mdseries,
  title=Definition \thetcbcounter: #2, label=#1,
}

\definecolor{tableheader}{RGB}{182,216,210}
\definecolor{tablerow}{RGB}{220,239,235}
\definecolor{tablebest}{RGB}{195,225,219}

\definecolor{pastelblue}{RGB}{185,195,235}
\definecolor{pastelbluebg}{RGB}{232,236,250}
\definecolor{pastelgreen}{RGB}{175,220,200}
\definecolor{pastelgreenbg}{RGB}{228,245,236}
\definecolor{pastelrose}{RGB}{230,185,185}
\definecolor{pastelrosebg}{RGB}{248,232,232}
\definecolor{promptgray}{RGB}{245,245,248}
\definecolor{promptframe}{RGB}{200,200,210}

\tcbset{promptboxstyle/.style={
  enhanced, breakable,
  coltitle=black, fonttitle=\bfseries\small,
  boxrule=0.6pt, arc=2pt,
  left=4pt, right=4pt, top=2pt, bottom=2pt,
  toptitle=2pt, bottomtitle=2pt,
  attach boxed title to top left={yshift=-2mm, xshift=4mm},
}}

\newtcolorbox{systempromptbox}[1][]{promptboxstyle, title={#1},
  colback=pastelbluebg, colframe=pastelblue,
  boxed title style={colback=pastelblue, colframe=pastelblue, arc=1.5pt,
                     boxrule=0pt}}
\newtcolorbox{userpromptbox}[1][]{promptboxstyle, title={#1},
  colback=pastelgreenbg, colframe=pastelgreen,
  boxed title style={colback=pastelgreen, colframe=pastelgreen, arc=1.5pt,
                     boxrule=0pt}}
\newtcolorbox{skillpromptbox}[1][]{promptboxstyle, title={#1},
  colback=pastelrosebg, colframe=pastelrose,
  boxed title style={colback=pastelrose, colframe=pastelrose, arc=1.5pt,
                     boxrule=0pt}}

\lstdefinestyle{promptstyle}{
  basicstyle=\ttfamily\scriptsize,
  breaklines=true, breakatwhitespace=false,
  columns=fullflexible, keepspaces=true,
  frame=none, xleftmargin=0pt,
  aboveskip=0pt, belowskip=0pt,
  literate={\ \ }{{\ \ }}2
           {—}{{\textemdash}}1
           {→}{{$\rightarrow$}}1
           {≈}{{$\approx$}}1,
}

\def\BibTeX{{\rm B\kern-.05em{\sc i\kern-.025em b}\kern-.08em
    T\kern-.1667em\lower.7ex\hbox{E}\kern-.125emX}}

\begin{document}


\title{\system{}: From Knowledge Repositories to Agent-Native Knowledge Infrastructure for Cyber Threat Intelligence}

\author{%
\IEEEauthorblockN{%
Yutong Cheng\textsuperscript{1,*},
Hao Fu\textsuperscript{1,*},
Changze Li\textsuperscript{1},
Qian Cui\textsuperscript{2},
Wei Ding\textsuperscript{2},
Lingzhi Wang\textsuperscript{3},
Yan Chen\textsuperscript{3}, and
Peng Gao\textsuperscript{1}
}
\IEEEauthorblockA{%
\textsuperscript{1}\textit{Virginia Tech}, Blacksburg, VA, USA \quad
\textsuperscript{2}\textit{Amazon}, USA \quad
\textsuperscript{3}\textit{Northwestern University}, Evanston, IL, USA
}
\IEEEauthorblockA{%
Email: \{yutongcheng, haofu26, changzeli, penggao\}@vt.edu,
\{cuiqia, dingwe\}@amazon.com,\\
LingzhiWang2025@u.northwestern.edu, ychen@northwestern.edu
}
\IEEEauthorblockA{%
\textsuperscript{*}These authors contributed equally to this work and share first authorship.
}
}

\maketitle

\begin{abstract}
Cyber threat intelligence (CTI) serve as important knowledge repositories in support of threat investigations, while existing research focuses on constructing repositories and representing threat knowledge, leaving dynamic, interactive queries to downstream applications such as diverse LLM-based agents.
Since agents increasingly adopted in threat investigations span heterogeneous CTI sources and adapts evidence-specific operations, we argue that CTI research should move beyond static knowledge construction toward a paradigm unifying data structures and operations.
Accordingly, we propose \system{}, an agent-native CTI infrastructure integrating heterogeneous CTI sources within a shared knowledge base and provides a unified operational interface for agents.
To achieve unified representation, \system{} constructs deterministic security entity concepts and entity relations as the foundational CTI storage, while complementing these ontological threat records by linking entity evidence across diverse CTI reports to support heterogeneous data storage.
Building on this foundation, \system{} provides a hybrid query interface that combines CTI ontology traversal with semantic similarity retrieval.
We package \system{} as modular infrastructure for agents, with extensible typed tools for CTI operations, and task-specific analytical skills.
Our nine-task CTI evaluation shows that \code{gpt-5.4} and \code{claude-haiku-4-5} agents using \system{} improve F1 by 22.3 and 26.0 points, respectively.
Furthermore, both Claude models make approximately half as many tool calls with \system{} as with their respective baselines, while 1{,}168 end-to-end investigations using \code{gpt-5.4} average only 2.6 US cents each.
Taken together, these results demonstrate that building agent-native knowledge infrastructure from CTI repositories enables more accurate and efficient investigations.
\end{abstract}



\section{Introduction}
\label{sec:intro}
Cyber threat intelligence (CTI) enables proactive defense, as security analysts rely on broad threat knowledge to conduct investigations~\cite{bouwman2020different}.
In practice, CTI spans heterogeneous sources and formats, ranging from unstructured threat reports and curated feeds to structured APIs~\cite{cticonnect}.
Prior research has primarily focused on extracting and organizing CTI~\cite{buechel2025ttpextraction}, for example by collecting indicators of compromise (IoCs) from public reports~\cite{liao2016iace} or constructing cybersecurity knowledge graphs~\cite{cheng2025ctinexus}.
However, real-world threat investigations involve sophisticated operations, such as linking suspicious domains and infrastructure indicators to exploited vulnerabilities, observed tactics, and attributed threat campaigns~\cite{zhu2018chainsmith}.
While some efforts utilize uniformly structured CTI sources for multi-source correlation, predefined intelligence and query operations expose limited context~\cite{li2019tealeaves}.
As agents are increasingly deployed for automated threat investigations to access broad CTI knowledge through tool invocation and retrieval augmentation~\cite{wu2026excytinbench,schick2023toolformer,lewis2020rag,cticonnect}, dynamic data access and interactive querying remain coupled to individual agents, which rely on custom parsing logic for heterogeneous CTI repositories.
However, querying across diverse heterogeneous data sources relies on intricate access logic, particularly for cross-source traversal and correlation~\cite{cticonnect}, while evolving sources constrain evidence coverage to each individual deployment~\cite{bouwman2020different,ethembabaoglu2026aptdisagree}.
Achieving efficient threat investigations across heterogeneous CTI sources and diverse agents remains an open problem.

Since agent-based threat investigations span heterogeneous data sources and associated workflows while interleaving data queries, reasoning, and tool calls across diverse models and APIs~\cite{yao2023react,wu2026excytinbench,wang2026mcpbench}, such fragmented operations are inherently inefficient, inconsistent and error-prone.
We argue that CTI research should move beyond static knowledge extraction, passive representation, and fragmented tool invocations toward an \textbf{agent-native knowledge infrastructure} that integrates data structures with composable operations and exposes a unified CTI access interface as a service to diverse agents.
However, establishing such an infrastructure with unified data storage and access interfaces presents three key technical challenges.


\noindent(i)~\emph{Semantic disambiguation.} Heterogeneous CTI sources assign inconsistent names to the same threat entity or concept, preventing consistent query results during investigations. For example, some sources apply \emph{Lazarus} as an umbrella term, whereas others track subsets as distinct groups, including \emph{APT38}~\cite{mitre-apt38}. Conflating these labels risks misattributing a broad campaign to a narrower threat group.

\noindent(ii)~\emph{Semantic alignment.}
Conducting investigations over unstructured CTI synthesizes evidence across multiple sources, yet existing queries struggle to preserve fine-grained provenance for inferred conclusions~\cite{slobodkin2024attributefirst}. For instance, one report may recommend monitoring a behavior, while another reports observing it. Without the supporting passage, synthesis can mistake that recommendation for a second independent observation.

\noindent(iii)~\emph{Semantic querying.} Existing retrieval-augmented generation (RAG) methods~\cite{lewis2020rag} rely on textual similarity and fall short on structured, multi-hop queries across heterogeneous CTI sources~\cite{sun2024tog,cticonnect}. For example, a report mentions a CVE while the question asks for its recorded CWE. Although text search retrieves topically similar weaknesses whose descriptions resemble the context, identifying the documented CWE depends on following the official relational link from that specific CVE~\cite{cticonnect}.

\noindent\textbf{Our work.} To address these challenges, we present \system{}, an agent-native knowledge infrastructure for cyber threat intelligence that consolidates heterogeneous CTI into a shared knowledge base and exposes a unified operational interface to downstream agents. \system{} is organized into a bottom-up, three-layer architecture:

To address the semantic disambiguation challenge, \system{} introduces the \emph{entity foundation layer}, which organizes full-text entity records from authoritative CTI repositories into a security ontology. As the foundational layer, it normalizes source records into a deterministic schema, retains canonical identifiers and official cross-references, and supplies authoritative canonical entities to anchor mention resolution.

To address evidence alignment, \system{} integrates an \emph{evidence alignment layer} that links co-referent entity mentions across heterogeneous CTI sources at character-level precision and augments the underlying ontology with report evidence. Cross-source entity alignment allows agents to synthesize evidence for the same entity across sources and verify findings against original text. By aligning evidence across sources, this layer allows agents to synthesize distributed entity findings and verify conclusions against raw source text.

To address the third challenge, \system{} incorporates a \emph{retrieval layer} that exposes a hybrid query interface over the shared knowledge base, combining ontology-guided traversal with semantic and lexical retrieval. The hybrid query interface allows agents to follow structured relationships, retrieve contextual evidence, and iteratively refine queries as new leads emerge during an investigation.

In summary, we make three key technical contributions.
\begin{itemize}[leftmargin=*,topsep=2pt,itemsep=2pt,parsep=0pt]
\item We propose a unified CTI storage approach consolidating heterogeneous CTI repositories into a shared knowledge base through ontology-grounded report storage.
\item We propose an alignment method for conflicting text across massive, heterogeneous CTI sources, establishing exact links between inferred conclusions and supporting source evidence.
\item We propose a unified query view that decouples domain-specific CTI data processing from downstream agent execution, allowing diverse agents to reuse common investigative operations and shared knowledge base.
\end{itemize}

To support broad adoption across existing agent frameworks, we package \system{} as a modular infrastructure offering seven typed tools for CTI operations and three task-specific analytical skills. We evaluate \system{} on nine tasks of CTIConnect~\cite{cticonnect} using mini-swe-agent~\cite{yang2024sweagent} under two configurations, where a flat-file baseline operates directly over raw CTI files, whereas \system{} serves the agent through unified query interface. Both settings share identical LLM, agent loop, step budget, and temperature parameters. Our evaluation shows these primary results.
\begin{enumerate}[label=\arabic*),leftmargin=*,topsep=2pt,itemsep=2pt,parsep=0pt]
\item \textbf{\system{} broadly improves agent investigation accuracy.} Across all evaluated models, F1 scores increase by $17.5$ to $26.0$ points. Specifically, overall F1 rises from $0.609$ to $0.832$ on \code{gpt-5.4} and from $0.483$ to $0.743$ on \code{claude-haiku-4-5} (Table~\ref{tab:main}).
\item \textbf{\system{} allows smaller models to outperform stronger frontier baselines.} Equipped with \system{}, \code{claude-haiku-4-5} achieves an overall F1 of $0.743$, outperforming \code{gpt-5.4} with flat-file access ($0.609$) by $13.4$ F1 points.
\item \textbf{\system{} substantially reduces agent tool-call overhead.} Both Claude models require less than half as many tool calls per investigation, with with average calls dropping from $12.33$ to $5.53$ on \code{claude-haiku-4-5} and from $9.82$ to $4.59$ on \code{claude-sonnet-4-6} (Table~\ref{tab:effort}).
\item \textbf{\system{} generalizes across models.} Across four models from two vendors, \system{} improves accuracy on $31$ out of $36$ model--task pairs across four models from two providers, including all $16$ entity-linking pairs, where average F1 increases by $41.5$ points~(Table~\ref{tab:main}).
\item \textbf{Tools and skills reinforce each other.} With \code{gpt-5.4}, overall F1 improves by $11.1$ points with tools alone and by $6.5$ points with skills alone, whereas combining both raises F1 by $22.3$ points, exceeding their sum by $4.7$ points (Table~\ref{tab:ablation}).
\item \textbf{\system{} generalizes to other CTI tasks.} On $20$ CTIBench examples, F1 increases from $0.130$ to $0.540$ for weakness mapping and from $0.339$ to $0.568$ for technique extraction (\S\ref{sec:eval:downstream}).
\end{enumerate}

The evaluation results show that \system{} enables more accurate and efficient threat investigations. While scaling model capabilities or refining agent harnesses can improve investigative performance, building a unified knowledge infrastructure is more economical and matters even more than deploying stronger models.

\section{Background and Motivation}
\label{sec:background}

\subsection{Cyber Threat Intelligence}
\label{sec:background:eco}
Cyber threat intelligence (CTI) combines threat observations with reference knowledge to interpret observed activity. Vendor reports provide text artifacts such as incident write-ups, actor profiles, and malware analyses, while curated feeds distribute concrete indicators including suspicious domains and file hashes~\cite{li2019tealeaves}. Alongside these sources, structured knowledge bases catalog vulnerabilities (CVE), weakness classes (CWE), attack patterns (CAPEC), and adversary techniques (MITRE ATT\&CK~\cite{mitre-attack}). Many of these records contain explicit cross-references, such that a CVE entry points to an underlying CWE weakness and a CAPEC entry links to both related weaknesses and ATT\&CK techniques, while ATT\&CK organizes techniques across fourteen tactics and connects sub-techniques to parent techniques. Together, unstructured reports and structured references provide investigators with evidence across complementary levels of granularity.

A single investigation often spans multiple forms of evidence. Starting from a suspicious domain, an analyst may consult vendor reports to identify associated adversary activity, and extract a vulnerability identifier to retrieve the corresponding weakness record, before returning to the reports to verify whether that weakness explains the observed behavior. In practice, vendors often track the same activity under different aliases, while reports frequently describe behaviors without naming the corresponding reference entries. Similarly, although feed lookups locate isolated indicators, situating an indicator within a broader campaign depends on corroborating evidence across sources. Effective cross-source investigation therefore demands linking heterogeneous records while retaining the investigative context needed to guide subsequent operations.

\subsection{Related Work}
\label{sec:related}

\noindent\textbf{CTI Knowledge Representation.}
Research on CTI reports has evolved from extracting indicators with surrounding context~\cite{liao2016iace} to extracting actions and attack patterns~\cite{husari2017ttpdrill,alam2023ladder}, as well as modeling campaign stages and attack behavior~\cite{zhu2018chainsmith,satvat2021extractor}. More recently, AttacKG constructs technique knowledge graphs, while CTINexus uses LLMs to extract entities and relations, align entities, and predict missing links~\cite{li2022attackg,cheng2025ctinexus}. Complementing report-based extraction, BRON complements links structured reference sources~\cite{hemberg2021bron}. Broader work on unstructured data further introduces methods for constructing and validating knowledge bases such as DeepDive and Fonduer, alongside frameworks for semantic operators and document pipelines, including LOTUS and DocETL~\cite{zhang2017deepdive,wu2018fonduer,patel2025lotus,shankar2025docetl}.
However, focusing solely on repository construction leaves downstream agent applications to implement their own cross-source alias resolution and evidence lookup.

\noindent\textbf{Cross-Source CTI Retrieval.}
Conventional RAG ranks passages by query relevance~\cite{lewis2020rag}, whereas RAPTOR organizes text hierarchically and systems such as GraphRAG and HippoRAG construct derived graphs to connect evidence across documents~\cite{sarthi2024raptor,edge2024graphrag,gutierrez2024hipporag}. Lexical and dense retrievers recover complementary match types, which rank fusion combines into unified rankings~\cite{robertson2009bm25,cormack2009rrf}. On structured sources, StructGPT and Think-on-Graph query relational schemas or traverse graph edges directly~\cite{jiang2023structgpt,sun2024tog}. Within the security domain, CTIBench evaluates LLMs on CTI tasks~\cite{alam2024ctibench}, and CTIConnect evaluates retrieval across reference records and vendor reports on entity linking, entity attribution, and multi-document synthesis~\cite{cticonnect}. Solving such tasks involves both locating relevant narrative passages and following official cross-references. However, decoupling passage ranking from graph traversal leaves the application without a direct mapping from text matches to canonical records prior to traversal.

\noindent\textbf{Agent Access to CTI Knowledge.}\label{sec:related:agents}
Frameworks such as ReAct and Toolformer interleave reasoning with tool invocations~\cite{yao2023react,schick2023toolformer}, while Search-R1 and WebThinker leverage intermediate findings to guide subsequent queries~\cite{jin2025searchr1,li2025webthinker}. Building on these capabilities, general-purpose harnesses, tool protocols, and modular skills offer standardized mechanisms for invoking external operations~\cite{yang2024sweagent,miniswe,mcp2024,anthropic-skills}. However, the effectiveness of iterative access depends on whether an agent can resolve canonical records, traverse official relationships, and revisit supporting passages. Relying solely on a general-purpose harness over raw file or feed interfaces leaves each deployment to repeatedly implement cross-source access and evidence verification.

\subsection{Challenges and Motivation}
Existing work on CTI and agent tooling addresses individual stages, from entity extraction to tool invocation, yet no unified framework currently supports all these stages as a unified single service. In practice, realizing such an infrastructure faces three core challenges.

\emph{(i) Semantic disambiguation.}
Entity extraction offer a foundation for identifying mentions across reports~\cite{liao2016iace,cheng2025ctinexus}. At query time, an agent starting from a vendor-specific name must locate relevant reports filed under other aliases while distinguishing closely related entities before aggregating evidence. In practice, surface-form matching misses aliases with divergent wording, whereas overly aggressive merging conflates separately tracked intrusion groups or malware families~\cite{christophides2020entity,cticonnect}. Resolving both failure modes calls for canonical entity records that preserve source-specific mentions and support retrieval via either an alias or a reference identifier.

\emph{(ii) Semantic alignment.}
Prior work on provenance and knowledge-base construction enables extracted facts to be verified against source material~\cite{zhang2017deepdive,wu2018fonduer,patel2025lotus}. Within CTI, campaign analysis and multi-document synthesis involve combining evidence across multiple reports and reference sources~\cite{zhu2018chainsmith,cticonnect}. During a multi-step investigation, downstream reasoning frequently builds on a claim derived from an earlier passage. However, a document-level link identifies only the enclosing report, leaving the agent to re-locate the supporting span within lengthy text. Reliable multi-hop verification therefore calls for exposing the extracted passage, source attribution, and exact character offsets so the agent can inspect the evidence before incorporating a finding into a subsequent query or conclusion.

\emph{(iii) Semantic querying.}
Hybrid retrieval recovers both paraphrased descriptions and exact terms, whereas graph traversal follows relationships already encoded in structured data~\cite{robertson2009bm25,cormack2009rrf,jiang2023structgpt,sun2024tog}. Threat analysis often combines both operations in sequence, such as searching for a paraphrased vulnerability description, retrieving a candidate CVE record, and following its recorded link to the corresponding CWE entry. However, similarity-based retrieval can return a related weakness even when the official mapping points elsewhere~\cite{lewis2020rag,cticonnect}. The interface therefore calls for exposing text search and relationship traversal as distinct operations over a shared set of records.

The challenges of evidence-driven threat analysis motivate a unified, agent-native knowledge infrastructure, since fragmented CTI repositories remain difficult to reuse across diverse agent workflows. To address these challenges, \system{} combines a shared entity representation, provenance links from entities to supporting passages in source documents, and a hybrid retrieval interface.

\section{System Design}
\label{sec:design}

\begin{figure*}[t]
\centering
\includegraphics[width=\linewidth]{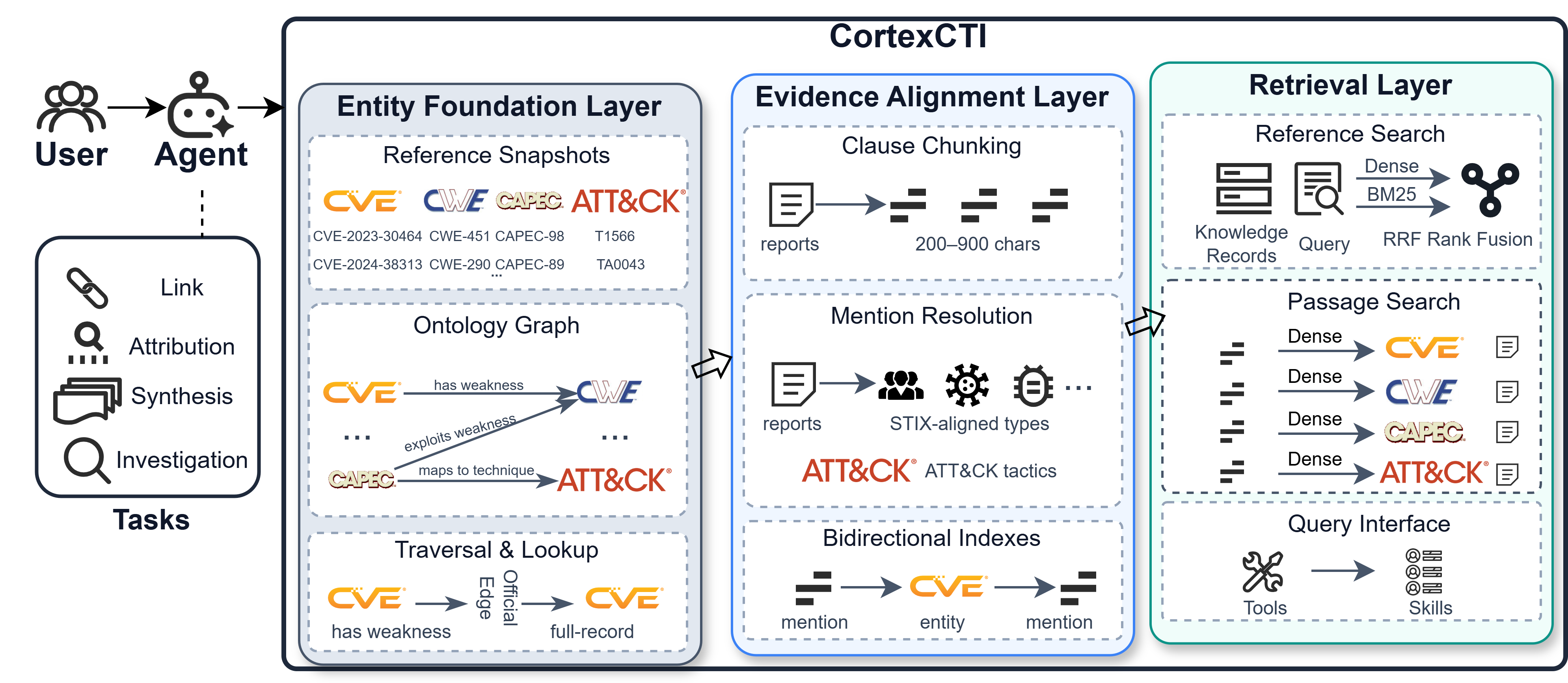}
\caption{In \system{}, the entity foundation layer validates official reference records and cross-taxonomy links, the evidence alignment layer resolves report mentions while preserving character-level source spans, and the retrieval layer indexes structured records and grounded passages, allowing diverse agent harnesses to invoke a shared typed tool interface for search, lookup, and traversal.}
\vspace{-0.3cm}
\label{fig:overview}
\end{figure*}

\subsection{Overview}
\label{sec:design:overview}

\system{} allows agents to carry evidence seamlessly between unstructured vendor reports and structured reference records by integrating report corpora and their metadata with CVE, CWE, CAPEC, and MITRE ATT\&CK entries within a shared representation. Formally, let $R$ denote the set of vendor reports and associated metadata, $K$ denote the structured reference records, and $C=R\cup K$ denote the combined corpus, with $\mathsf{S}(C)$ representing its unified schema. To accommodate diverse CTI sources and agents, this design follows three properties.

\noindent\emph{(i) Consistent entity identity.}
Mentions with evidence of equivalence share a canonical report entity, while lookups by an authoritative identifier resolve to the corresponding structured reference record.

\noindent\emph{(ii) Traceable evidence.}
Each evidence passage records its source report and exact character offsets, with stored passage text matching the source at those offsets.

\noindent\emph{(iii) Reusable query results.}
Search and lookup operations return identifiers or passage references that follow-up calls can use to address the same evidence directly.

\stitle{System Architecture.}\phantomsection
\label{sec:design:build}
The three layers in Fig.~\ref{fig:overview} realize these properties across the pipeline from corpus construction to query execution.

To anchor investigations in authoritative evidence, the \emph{Entity Foundation Layer} constructs the ontology graph $G=(V,E)$ from released knowledge-base snapshots, where nodes in $V$ store complete records under canonical identifiers and edges in $E$ encode official cross-references. Endpoint validation further verifies that every traversable link resolves to an existing record within the snapshots.

To connect vendor evidence to these reference records and across aliases, the \emph{Evidence Alignment Layer} resolves report mentions into canonical entities $\mathcal{E}$ and stores supporting passages in $X$. An inverted index $\Pi:\mathcal{E}\to 2^X$ maps each entity to its supporting passages, while validation verifies every passage against the original report text at its recorded character offsets. Grounded mentions further link to $G$ through verified identifiers, and the build enforces $\mathrm{ids}_K(\mathsf{S}(C))\subseteq\mathrm{ids}(K)$, where $\mathrm{ids}_K(\mathsf{S}(C))$ denotes the authoritative identifiers referenced by the representation and $\mathrm{ids}(K)$ the canonical identifiers present in $K$. An agent can therefore gather evidence under a single entity identifier and verify every passage directly in its source.

To make these records and passages accessible from text queries, the \emph{Retrieval Layer} constructs search indexes $\mathcal{I}$ spanning both underlying layers. Reference search fuses dense and lexical rankings to return canonical identifiers, whereas dense passage search returns text spans paired with source offsets and associated entities. These returned references serve directly as inputs for follow-up lookups and graph traversal.

The resulting knowledge base operates as a reusable data store. Much like applications sharing a relational database, distinct agents can query the same validated artifacts through a common interface. In \system{}, these artifacts comprise the ontology graph, report passages, canonical entities, and search indexes, denoted as $\mathsf{S}(C)=(G,X,\mathcal{E},\Pi,\mathcal{I})$. Offline construction links report evidence to the ontology via verified identifiers and indexes it for persistent reuse. Consequently, agents can retrieve and verify evidence across investigations without re-running extraction over raw reports or retraining the underlying model, while each harness independently selects its model, step budget, and call sequence.

\begin{table}[t]
\papertablestyle
\caption{Corpus build summary covering ontology nodes and edges, report
artifacts, one-time cost, and validation checks.}
\label{tab:scaffold}
\begin{tabularx}{\linewidth}{L r L r}
\toprule
\multicolumn{2}{l}{\textbf{Ontology nodes}} &
\multicolumn{2}{l}{\textbf{Typed edges}} \\
\midrule
CVE     & 3{,}011 & \code{has\_weakness}       & 3{,}290 \\
CWE     & 1{,}342 & \code{exploits\_weakness}  & 1{,}214 \\
CAPEC   &    615  & \code{in\_tactic}          & 1{,}076 \\
ATT\&CK & 1{,}076 & \code{child\_of}           &    533 \\
\textit{Total} & \textit{6{,}044} & \code{sub\_technique\_of} & 518 \\
\multicolumn{2}{l}{\textbf{Aligned evidence}} &
  \code{maps\_to\_technique} & 272 \\
Vendor reports & 321 & CAPEC ordering & 194 \\
Span-grounded chunks & 688 & \textit{Total} & \textit{7{,}097} \\
Canonical entities & 4{,}868 & & \\
\midrule
\multicolumn{2}{l}{\textbf{Build cost}} &
\multicolumn{2}{l}{\textbf{Validation}} \\
\midrule
Build tokens & 1.42M & Fabricated identifiers & 0 \\
Build cost & \$1.86 & Orphan edges & 0 \\
 & & Span violations & 0 \\
\bottomrule
\end{tabularx}

\end{table}

\begin{table*}[t]
\papertablestyle
\caption{Seven typed tools for searching, looking up, and connecting
evidence, with the cost category of each operation.}
\label{tab:tools}
\begin{tabularx}{\linewidth}{>{\raggedright\arraybackslash}p{0.205\linewidth} L l}
\toprule
\textbf{Tool} & \textbf{Capability} & \textbf{Cost} \\
\midrule
\code{resolve\_entity} & name/alias/id $\to$ canonical entities with
  cross-vendor alias lists and grounded external ids; prescribed first call
  & cheap \\
\code{get\_entity} & full record of one ontology node or report entity
  & cheap \\
\code{ontology\_neighbors} & traverse official cross-reference edges
  (\code{has\_weakness}, \code{exploits\_weakness},
  \code{maps\_to\_technique}, \dots); deterministic
  & cheap \\
\code{search\_kb} & hybrid dense+BM25 search over one KB, with
  \code{must\_terms} conjunctive filter and pool-size feedback
  & moderate \\
\code{search\_chunks} & dense search over report chunks; returns text,
  associated entities, and TTPs for follow-up graph queries
  & moderate \\
\code{chunks\_mentioning} & all chunks of every report where an entity
  appears, with full text---the cross-vendor collection primitive
  & cheap \\
\code{read\_report} & full text of one report by document id
  & cheap \\
\bottomrule
\end{tabularx}

\end{table*}

\subsection{Entity Foundation Layer}
\label{sec:design:onto}

\begin{figure*}[t]
\centering
\includegraphics[width=\linewidth]{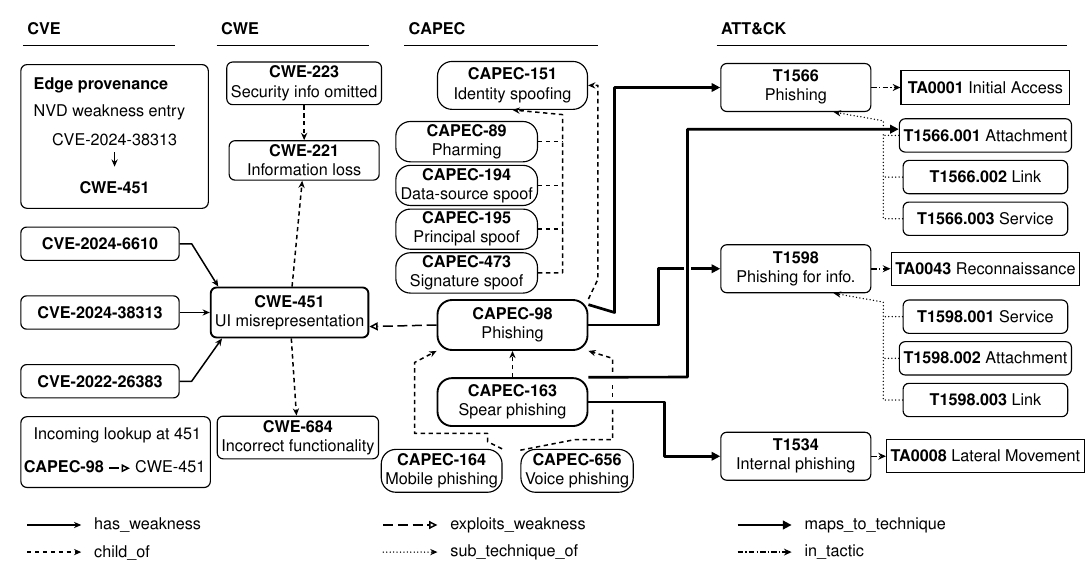}
\caption{Typed links across CVE, CWE, CAPEC, and ATT\&CK within a 28-record ontology excerpt, where the NVD record for CVE-2024-38313 points to CWE-451 via \code{has\_weakness} and CAPEC-98 points to the same weakness via \code{exploits\_weakness}. An incoming lookup at CWE-451 recovers CAPEC-98 while preserving both the orientation and provenance of the underlying stored edge.}
\label{fig:foundation-layer}
\end{figure*}

\stitle{Canonical records and edges.}
Verifying a cross-taxonomy link requires access to the original record. To support this, the foundation layer instantiates records in $K$ as nodes in $V$, indexes them by canonical identifier, and retains the full source payloads. Each admitted cross-reference forms a directed edge $(u,\ell,v)$ between existing nodes, where $u$ denotes the source node, $v$ denotes the destination node, and $\ell$ denotes the relation type. Specifically, recorded CVE-to-CWE links take the label \code{has\_weakness}, whereas CAPEC-to-ATT\&CK mappings use \code{maps\_to\_technique}, and intra-source edges such as \code{child\_of} and \code{sub\_technique\_of} preserve hierarchies within individual taxonomies (Table~\ref{tab:scaffold}).

Traceability also depends on strict edge admission rules. The foundation layer parses pinned snapshots deterministically and admits an edge only when an upstream record explicitly declares the cross-reference and both endpoint identifiers resolve in $K$. Since originating record provides the authoritative evidence for verification while pinned snapshots hold underlying sources fixed across builds, every graph link remains auditable and reproducible.

\stitle{Typed traversal.}
Once links are validated, agents can filter neighbors by relation type and direction to answer targeted questions. Hierarchy edges link parent and child entries within a single reference source, whereas mapping edges bridge records serving distinct analytical roles. Given a node $u$ and relation type $\ell$, outgoing traversal yields $N_{\ell}^{+}(u)=\{v\in V : (u,\ell,v)\in E\}$.

Traversing \code{has\_weakness} edges from a CVE record retrieves upstream-declared CWE entries (Fig.~\ref{fig:foundation-layer}), while a subsequent full-record lookup at each destination supplies the weakness definition and surrounding context.

The same directed edge supports queries from either endpoint. For example, an incoming traversal on CWE-451 along \code{exploits\_weakness} recovers CAPEC-98, whereas an outgoing traversal from a CVE follows  \code{has\_weakness} links. Query direction simply specifies which endpoint the caller holds, while stored edges preserve original semantics in both directions and keep originating records accessible for cross-reference verification.

\subsection{Evidence Alignment Layer}
\label{sec:design:report}
\label{app:build:steps}
\label{app:resolve}

\begin{figure*}[t]
\centering
\includegraphics[width=0.86\linewidth]{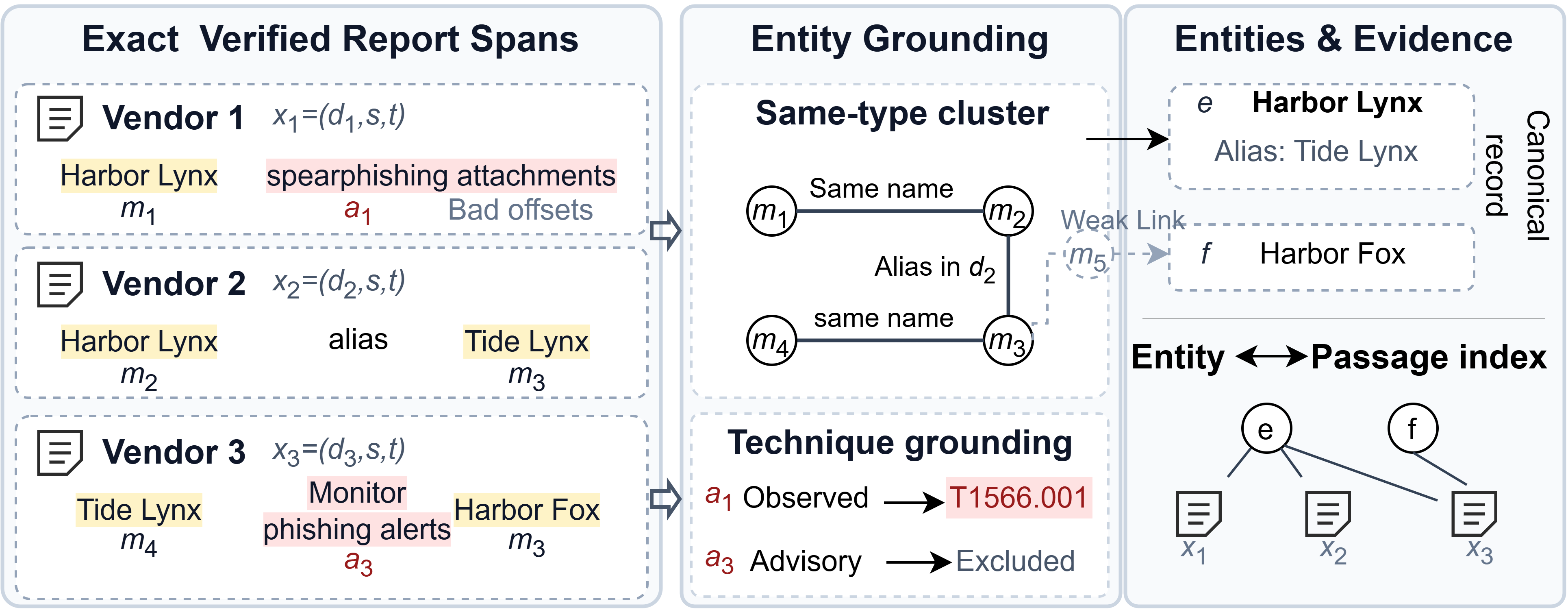}
\caption{Evidence alignment across three illustrative reports, where verified spans preserve the exact text naming an entity or describing observed behavior. Supported aliases consolidate mentions under a canonical entity whose identifier retrieves associated passages across reports, while grounded technique identifiers link behavioral spans directly to their corresponding ATT\&CK records.}
\label{fig:entity-resolution}
\end{figure*}

\stitle{Span-grounded extraction.}
Connecting the reference graph to vendor reports depends on preserving the local context of each observation. To achieve this, the alignment layer segments reports at clause boundaries and packs contiguous clauses into chunks of 200 to 900 characters. Each chunk $x\in X$ records a provenance tuple $(d,s,t)$, where $d$ identifies the report and $s$ and $t$ denote the start and end character offsets in the frozen source text $T_d$$T_d$. During validation, the pipeline enforces the invariant
\[
  \operatorname{text}(x)=\operatorname{slice}(T_d,s,t),
\]
where $\operatorname{text}(x)$ is the stored chunk text and $\operatorname{slice}$ extracts the substring at those offsets. Across the evaluation corpus, this partitioning yields 688 chunks, each linking a retrieved passage back to its surrounding context in the full report.

Cross-source resolution begins with extracting typed mentions from these passages. Building on CTINexus~\cite{cheng2025ctinexus}, an LLM extractor proposes entity mentions and within-chunk coreferences across eight STIX-aligned types, covering threat actors, malware, tools, techniques, vulnerabilities, campaigns, identities, and indicators. In parallel, deterministic patterns capture structured identifiers including CVE and ATT\&CK IDs, file hashes, and IP addresses. The build retains a candidate mention only when the cited span matches reported surface text, providing the downstream resolver with verifiable ground truth for every kept mention.

Observed adversary behavior must likewise link to the reference graph. A dedicated technique extractor maps verbatim behavioral descriptions to candidate ATT\&CK techniques while filtering out defensive recommendations. The build validates each predicted identifier against the pinned ATT\&CK snapshot and pairs the supporting passage with the proposed grounding. Agents can thus verify each proposed mapping directly against its source passage, while official taxonomy relationships remain isolated in $G$ apart from report-derived groundings.

\stitle{Entity resolution and evidence access.}
Grouping extracted mentions allows agents to aggregate evidence across vendor-specific aliases. The build applies union-find over mentions sharing the same entity type, checking in order for a shared grounded identifier, a shared ATT\&CK technique grounding, and a matching normalized name. Within-chunk coreference and explicit alias statements trigger additional merges whenever the source span verifies equivalence (Fig.~\ref{fig:entity-resolution}), whereas weak \emph{possibly-same} links remain outside the clusters to prevent uncertain equivalences from propagating. Each resulting cluster forms a canonical entity, yielding 4{,}868 entities that retain vendor-attributed aliases alongside reference identifiers for grounded records.

Once mentions are clustered, agents must traverse between entities and their supporting evidence in either direction. Let $\mathcal{M}_e$ denote the set of mentions assigned to canonical entity $e$ and let $\operatorname{chunk}(m)$ denote the source chunk containing mention $m$. The build constructs the forward and reverse indexes
\[
\begin{aligned}
  \Pi(e) &= \{\,\operatorname{chunk}(m):m\in\mathcal{M}_e\,\},\\
  \Gamma(x) &= \{\,e\in\mathcal{E}:x\in\Pi(e)\,\}.
\end{aligned}
\]
The forward index $\Pi(e)$ retrieves supporting passages by entity identifier, while the reverse index $\Gamma(x)$ identifies all entities appearing in a retrieved passage, with vendor-attributed aliases linking the collected evidence back to the original report names.

\subsection{Retrieval Layer}
\label{sec:design:dense}

\begin{figure*}[t]
\centering
\includegraphics[width=0.86\linewidth]{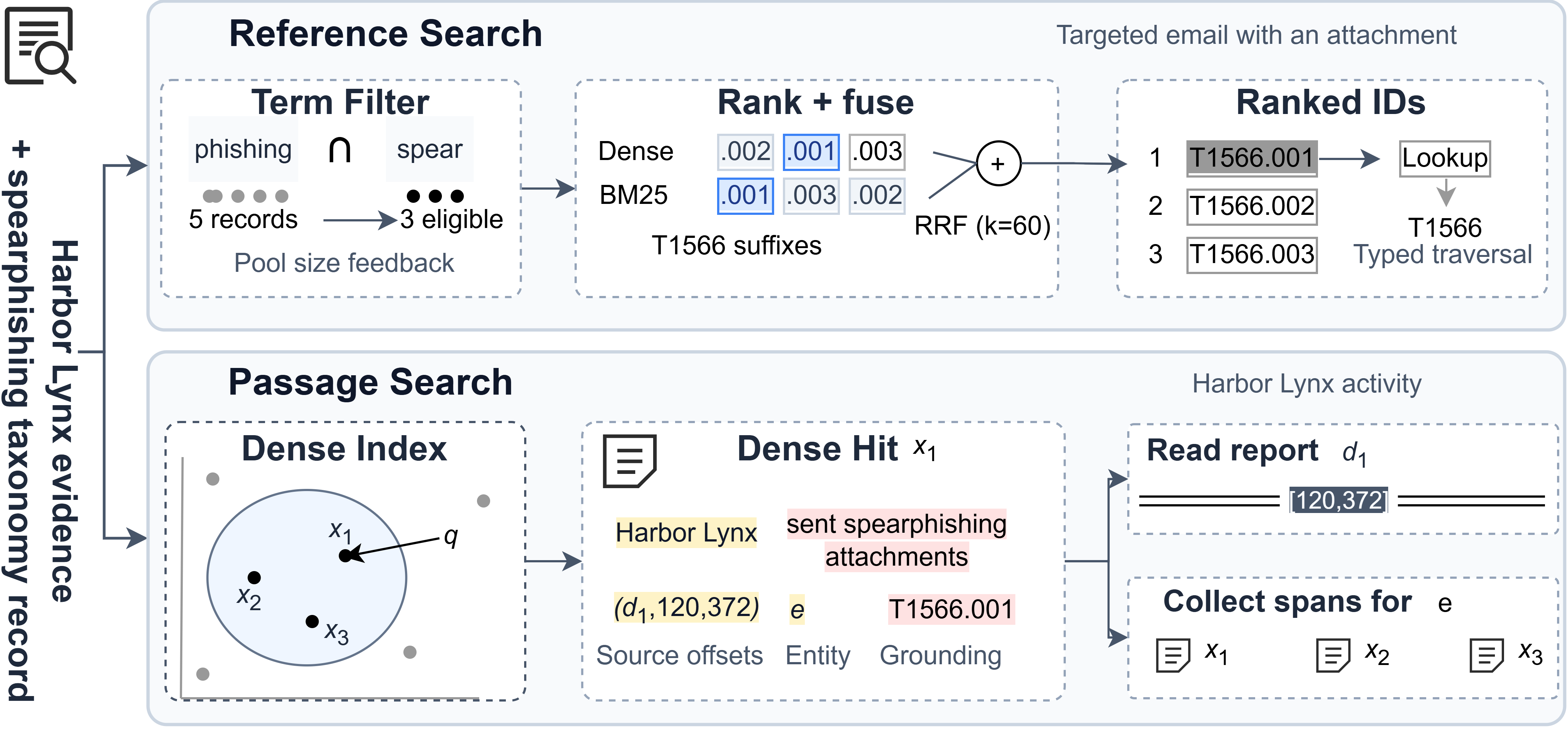}
\caption{Reference and passage retrieval for an illustrative investigation, where term filtering and rank fusion (rank offset $k=60$) select candidate reference records while source offsets and entity identifiers anchor retrieved passages. Returned canonical identifiers allow the agent to proceed directly to record lookup, graph traversal, or cross-report evidence collection.}
\label{fig:retrieval-flow}
\end{figure*}

\stitle{Hybrid search.}
When canonical identifiers are unknown, agents can query using a natural-language description of the target record or passage. Using \code{text-embedding-3-large} with disk-cached embeddings, the build constructs a dense index over report chunks alongside a separate index for each reference knowledge base. Dense retrieval surfaces evidence phrased differently from the query, with the report index locating passages for verification and the reference indexes identifying records for full-text inspection or graph traversal.

Exact term matches provide a complementary signal for ranking reference records. \system{} maintains a BM25 index for each knowledge base and combines the lexical and dense rankings via reciprocal-rank fusion~\cite{robertson2009bm25,cormack2009rrf}. Letting $\mathcal{L}(v)$ denote the set of ranked lists containing candidate record $v$, the fused score is computed as
\[
  \operatorname{RRF}(v)=
  \sum_{L\in\mathcal{L}(v)}\frac{1}{k+\operatorname{rank}_{L}(v)},
\]
where $k$ is the rank offset and $\operatorname{rank}_{L}(v)$ is the position of $v$ in list $L$, allowing records matched by both signals to accumulate credit.

\stitle{Refining and reusing search results.}
Broad candidate sets hinder individual record inspection, so reference search accepts a conjunctive term filter to narrow the search space. Let $V_j$ denote the records in the selected reference knowledge base $j$ and let $T$ denote the set of terms supplied in \code{must\_terms}. The eligible pool is defined as
\[
  V_j(T)=\{v\in V_j:\ \forall t\in T,\ \operatorname{match}(v,t)\},
\]
where $\operatorname{match}$ is the term predicate evaluated by the filter. Each response reports $|V_j(T)|$, allowing the agent to add a discriminative term when the pool remains broad or revise a term when the pool is empty (Fig.~\ref{fig:retrieval-flow}), while full-record lookup allows shortlisted candidates to be verified before citation. In the illustrative reference search in Fig.~\ref{fig:retrieval-flow}, pairing \emph{spear} with \emph{phishing} shrinks the eligible pool from five records to three. T1566.001 then ranks second in the dense list and first in the BM25 list, placing first after rank fusion so the agent can inspect its official definition via the returned identifier and follow its parent link whenever the broader technique is relevant.

Report search similarly supports follow-up queries by returning passage text, associated entities, and ATT\&CK groundings for verification or expansion. Entity identifiers serve as keys for retrieving additional supporting passages through $\Pi$, whereas reference identifiers serve as keys for full-record lookup and traversal of official relationships in $G$. The Harbor Lynx example in Fig.~\ref{fig:retrieval-flow} illustrates both handoffs. Passage search returns a report identifier for \code{read\_report} alongside a canonical actor identifier for \code{chunks\_mentioning}, allowing the agent to recover surrounding context and gather corroborating evidence across other reports, while the candidate technique identifier passes to \code{get\_entity} to verify the official definition against observed behavior.

\stitle{Agent tools and procedural skills.}\phantomsection
\label{sec:design:agent-interface}
A common query interface allows diverse harnesses to reuse the validated knowledge base across investigation tasks. The retrieval layer exposes seven strongly typed tools (Table~\ref{tab:tools}) that resolve entity names, fetch full records, traverse official relationships, search knowledge bases and report chunks, gather passages by entity, and read complete reports. Each tool description specifies when to invoke it and indicates its execution cost category~\cite{anthropic-writing-tools}, steering agents to resolve canonical names before collecting evidence and to follow official links whenever a reference identifier is known.

Three task-family skills govern how agents compose these tools in recurring investigations, each injecting structured Markdown playbooks into user turn (Appendix~\ref{app:skills}). For entity linking, the playbook starts from a source record and traverses official cross-references. For attribution, it translates observed behaviors into standard taxonomy terminology and verifies candidate identifiers against reference records. For synthesis, it gathers relevant reports and consolidates findings across vendors. Finally, the playbooks standardize output formats and answer validation rules.

\section{Evaluation}
\label{sec:eval}
We use CTIConnect~\cite{cticonnect} to evaluate knowledge construction and multi-step investigation, and public downstream benchmarks to assess cross-task transfer.

\begin{enumerate}[label=\textbf{RQ\arabic*:},leftmargin=*,align=left,topsep=3pt,itemsep=3pt,parsep=0pt]
\item How accurately does \system{} construct and validate CTI knowledge from reference records and vendor reports?
\item How does \system{} affect investigation accuracy under a fixed agent harness across tasks and models?
\item How do tool usage patterns differ across agents, and how do cost, latency, and accuracy vary across question sets under \system{}?
\item How much do typed tools and procedural skills contribute, both individually and together, to investigation accuracy?
\item How well does structured CTI access transfer to external benchmarks, and how does evidence selection shape downstream security detection and other evidence-grounded tasks?
\end{enumerate}

\subsection{Experimental Setup}
\label{sec:eval:setup}

\looseness=-1\stitle{Benchmark and configurations.}
CTIConnect comprises 1{,}859 expert-verified questions spanning four reference knowledge bases and 321 multi-vendor report summaries across nine tasks organized into three families. Specifically, the four entity linking (EL) tasks identify reference entities and cross-taxonomy relationships, covering Root Cause Mapping (RCM), Weakness Instantiation Mapping (WIM), Attack Technique Derivation (ATD), and Exploitation Surface Discovery (ESD). The two entity attribution (EA) tasks map report descriptions to ATT\&CK techniques or CWE weaknesses, consisting of Attack Technique Attribution (ATA) and Vulnerability Catalog Attribution (VCA). The three multi-document synthesis (MDS) tasks aggregate evidence across reports into campaign summaries, actor profiles, or malware lineages, covering Campaign Storyline Construction (CSC), Threat Actor Profiling (TAP), and Malware Lineage Analysis (MLA). Using this corpus to evaluate multi-step investigative agents, we partition the benchmark into a \emph{main set} of 691 questions for controlled comparisons and a \emph{scale set} of 1{,}168 questions for cost and volume analysis. Prior to tuning, we randomly sampled 92 questions from the scale set for development, froze all prompts, tools, and skills once tuning concluded, and excluded those development items from reported scale-set accuracy.

For controlled comparisons, both configurations run the stock mini-swe-agent~\cite{miniswe} \code{DefaultAgent} over the identical corpus with matched agent loops, 20-step budgets, temperatures, and underlying models. Whereas the \emph{base agent} queries flat files via the harness's native \code{bash} tool, the \emph{\system{} agent} accesses the same underlying data through our seven typed tools and three skills. We evaluate two models from each provider, namely \code{gpt-5.4} and \code{gpt-5.4-mini} from OpenAI alongside \code{claude-sonnet-4-6} and \code{claude-haiku-4-5} from Anthropic.

All \system{} query runs in RQ2 through RQ4 operate over a single frozen build constructed from the four reference snapshots and 321 vendor report summaries, using \code{gpt-5.4-mini} for extraction under the operator prompts detailed in Appendix~\ref{app:prompts}.

\stitle{Metrics.}
We apply the benchmark's evaluation metrics identically across all configurations. Both EL and EA use \emph{identifier-normalized F1}, which normalizes case, prefixes, and sub-technique suffixes in the final answer so that \code{T1059.001} and \code{T1059} resolve to the same identifier. Comparing the predicted set against the reference set treats every extraneous identifier as a false positive and every omitted identifier as a false negative, meaning that outputting unresolved candidates penalizes precision even when the correct identifier is present.

For MDS, the benchmark's claim-level judge runs \code{gpt-5.4} with an identical prompt across configurations, while \emph{Overall} reports the unweighted mean across all nine tasks. In addition, we measure tool use by counting invocations per question from recorded execution trajectories, record wall-clock elapsed time per investigation, and compute model cost from provider-reported token usage at published list rates.

To validate the reliability of MDS scoring, two auditors with $3{+}$ years of professional CTI experience independently re-scored a 50-item MDS sample blind to the judge's verdicts. This audit yielded a Pearson correlation of 0.85 and a mean absolute difference of 0.06 per item, with mean scores of 0.705 and 0.713. While the 0.06 difference quantifies item-level disagreement, uncertainty in configuration-level averages further depends on how judgment errors covary across items and configurations, whereas source-reference validity is evaluated separately through the build's structural checks.

\subsection{RQ1: Build Validity and Extraction Accuracy}
\label{sec:eval:build}
\label{app:addbase:extraction}

\stitle{Structural validity and entity resolution.}
From the 321 vendor reports and four reference snapshots, the build materializes 6{,}044 ontology nodes, 7{,}097 official edges, 688 provenance-carrying chunks, and 4{,}868 canonical entities. During construction, every identifier is verified against the reference snapshots at write time, and the validator rejects any build containing unknown identifiers, orphan edges, or recorded spans that fail a byte-for-byte comparison against the frozen source. The evaluated build passed all three checks, consuming 1.42 million tokens at a total cost of\$1.86.

To evaluate whether canonical entities accurately group reports describing the same adversary, we hold out the benchmark's 50 report clusters strictly as evaluation labels. For each gold cluster $g$, we select the canonical entity $\varepsilon$ whose report set $R(\varepsilon)$ maximizes overlap with $g$, reporting coverage $=|R(\varepsilon)\cap g|/|g|$, purity $=|R(\varepsilon)\cap g|/|R(\varepsilon)|$, and their harmonic mean. \system{} achieves 0.900 coverage, 0.927 purity, and 0.913 F1 while reconstructing 27 clusters exactly, outperforming the benchmark's extraction graph at 0.840 coverage, 0.920 purity, and 0.878 F1 with 21 exact clusters. The primary gain stems from coverage, which improves by 0.060 alongside a 0.007 increase in purity, because our merge procedure restricts cluster membership to exact matches and span-verified aliases. By contrast, in the benchmark extraction graph, only three of the nine reports assigned to the best match for \code{APT42} belong to the true gold cluster.

\stitle{Extraction accuracy and errors.}
We further evaluate individual mention extraction and technique assignments. For mention extraction, we evaluate all six operators on a gold set of roughly 60 stratified, human-audited chunks spanning all eight entity types, where any type mismatch counts as both a false positive and a false negative. For TTP grounding, we strip explicit ATT\&CK identifiers from passages with known groundings and mix in roughly 40 vague or defender-side phrases that models should leave unassigned. In Table~\ref{tab:rq1}, \code{gpt-5.4} attains the highest F1 at 0.859 for entity extraction and 0.792 for TTP grounding, whereas our build model \code{gpt-5.4-mini} reaches 0.743 and 0.649. \code{gpt-5.4-mini} trails \code{gpt-5.4} by 0.138 in precision and 0.091 in recall on entity extraction, and by 0.138 in precision and 0.149 in recall on TTP grounding. This precision gap motivates examining which erroneous predictions still pass the build's structural checks.

\begin{table}[t]
\papertablestyle
\setlength{\tabcolsep}{2.75pt}
\caption{Entity extraction and technique grounding by build model.
Bold values indicate the highest score in each column. Underlining indicates the runner-up.}
\label{tab:rq1}
\begin{tabularx}{\columnwidth}{l *{6}{C}}
\toprule
 & \multicolumn{3}{c}{\textbf{Entity extraction}} & \multicolumn{3}{c}{\textbf{TTP grounding}} \\
\cmidrule(lr){2-4}\cmidrule(lr){5-7}
\textbf{Operator model} & Prec & Rec & F1 & Prec & Rec & F1 \\
\midrule
gpt-5.4            & \second{0.846} & \best{0.872} & \best{0.859} & \best{0.749} & \best{0.841} & \best{0.792} \\
gpt-5.4-mini       & 0.708 & \second{0.781} & 0.743 & 0.611 & 0.692 & 0.649 \\
gpt-5.4-nano       & 0.691 & 0.712 & 0.701 & 0.574 & 0.580 & 0.577 \\
claude-opus-4-1    & \best{0.849} & 0.748 & \second{0.795} & \second{0.727} & 0.735 & \second{0.731} \\
claude-sonnet-4-6  & 0.778 & 0.771 & 0.775 & 0.627 & \second{0.806} & 0.705 \\
claude-haiku-4-5   & 0.780 & 0.653 & 0.711 & 0.551 & 0.599 & 0.574 \\
\bottomrule
\end{tabularx}

\end{table}

To characterize the build model's failure modes, we manually categorize all of its false positives on the gold set, summarizing 155 entity extraction errors and 63 TTP-grounding errors in Table~\ref{tab:fpcauses}.

\begin{table}[htbp]
\papertablestyle
\caption{\rev{Distribution of \code{gpt-5.4-mini} false positives by cause within each layer.}}
\label{tab:fpcauses}
\begin{tabularx}{\linewidth}{l L c}
\toprule
\textbf{Layer} & \textbf{Cause} & \textbf{Share} \\
\midrule
Entity & wrong type & 38\% \\
Entity & span boundary & 29\% \\
Entity & spurious mention & 25\% \\
Entity & other & 8\% \\
TTP & sibling technique & 35\% \\
TTP & parent/child confusion & 19\% \\
TTP & defender behavior & 11\% \\
TTP & should have abstained & 24\% \\
TTP & other & 11\% \\
\bottomrule
\end{tabularx}
\end{table}

Entity type errors constitute most mention false positives, whereas confusions between sibling or parent and child techniques account for more than half of the grounding false positives, alongside passages that should have remained unassigned (Table~\ref{tab:fpcauses}). Such semantic errors pass build-time validation whenever both the character span and the referenced identifier are valid. An exact span confirms only that a surface string appears in a report, while a valid identifier confirms only that the target reference entry exists. By preserving the underlying passage alongside each grounding, the build ensures that agents retain direct access to the original evidence when verifying whether a report actually describes the assigned technique.

\subsection{RQ2: Accuracy Across Threat Analysis Tasks and Models}
\label{sec:eval:query}
\label{app:addbase:pipelines}
\label{app:addbase:failures}

\begin{table*}[t]
\papertablestyle
\caption{Main-set scores with flat-file access and the \system{} interface
under matched settings for each model.
Bold values indicate the higher score for each model and metric.}
\label{tab:main}
\begin{tabularx}{\linewidth}{l l *{4}{C} *{2}{C} *{3}{C} C}
\toprule
 & & \multicolumn{4}{c}{\textbf{EL}} & \multicolumn{2}{c}{\textbf{EA}} & \multicolumn{3}{c}{\textbf{MDS}} & \\
\cmidrule(lr){3-6}\cmidrule(lr){7-8}\cmidrule(lr){9-11}
\textbf{Model} & \textbf{Config} & RCM & WIM & ATD & ESD & ATA & VCA & CSC & TAP & MLA & \textbf{Overall} \\
\midrule
\multirow{2}{*}{gpt-5.4}
 & Base      & 0.660 & 0.680 & 0.230 & 0.460 & 0.546 & 0.442 & 0.849 & 0.830 & 0.788 & 0.609 \\
 & \system{} & \best{0.980} & \best{0.760} & \best{0.990} & \best{0.900} & \best{0.656} & \best{0.636} & \best{0.889} & \best{0.871} & \best{0.808} & \best{0.832} \\
\addlinespace[2pt]
\multirow{2}{*}{gpt-5.4-mini}
 & Base      & 0.540 & 0.420 & 0.380 & 0.450 & 0.460 & \best{0.701} & 0.779 & 0.776 & \best{0.699} & 0.578 \\
 & \system{} & \best{0.930} & \best{0.630} & \best{0.910} & \best{0.890} & \best{0.614} & 0.545 & \best{0.787} & \best{0.787} & 0.690 & \best{0.754} \\
\addlinespace[2pt]
\multirow{2}{*}{claude-sonnet-4-6}
 & Base      & 0.300 & 0.680 & 0.200 & 0.740 & \best{0.599} & 0.326 & 0.749 & 0.747 & \best{0.739} & 0.564 \\
 & \system{} & \best{0.910} & \best{0.820} & \best{0.780} & \best{0.990} & 0.543 & \best{0.510} & \best{0.800} & \best{0.873} & 0.696 & \best{0.769} \\
\addlinespace[2pt]
\multirow{2}{*}{claude-haiku-4-5}
 & Base      & 0.390 & 0.270 & 0.190 & 0.640 & 0.440 & 0.327 & 0.659 & 0.738 & \best{0.691} & 0.483 \\
 & \system{} & \best{0.900} & \best{0.580} & \best{0.990} & \best{0.910} & \best{0.499} & \best{0.496} & \best{0.792} & \best{0.827} & 0.690 & \best{0.743} \\
\bottomrule
\end{tabularx}
\end{table*}

\stitle{Main-set results.}
Replacing flat-file access with \system{} raises the overall score by roughly 0.18 to 0.26 across the four models. Table~\ref{tab:main} reports gains in 31 of the 36 model and task pairs, including all 16 EL pairs, with task-family averages across models increasing by 0.415 on EL, 0.082 on EA, and 0.039 on MDS.

These gains also invert the relative ranking across models. Equipped with \system{}, \code{claude-haiku-4-5} reaches 0.743 overall and \code{claude-sonnet-4-6} reaches 0.769, both surpassing the 0.609 attained by \code{gpt-5.4} over flat files. In fact, Haiku's 0.134 advantage over flat-file GPT-5.4 exceeds the 0.126 deficit by which it trails GPT-5.4 when both operate on flat files.

EL exhibits the largest gains because its three forward tasks allow the agent to resolve a source identifier and follow an official cross-reference. With \code{gpt-5.4}, F1 climbs from 0.66 to 0.98 on RCM, from 0.23 to 0.99 on ATD, and from 0.46 to 0.90 on ESD, while \code{claude-haiku-4-5} with \system{} similarly reaches 0.99 on ATD and 0.91 on ESD. WIM evaluates the inverse direction by querying the source of a recorded relation, where base-agent scores span 0.27 to 0.68 and \system{} raises them by 0.08 to 0.31 across all four models.

In EA, retrieved taxonomy entries act as candidate labels that the agent verifies against report text, lifting mean EA F1 by 0.06 to 0.15 on three models. On \code{gpt-5.4-mini}, however, a decline on VCA offsets the gain on ATA, and the base agent also scores higher on ATA with \code{claude-sonnet-4-6}. In MDS, scores improve on CSC and TAP across all four models, whereas on MLA both agents draw evidence from the same underlying reports and paired scores differ by at most 0.043. Unlike EL queries that can be answered directly from a recorded cross-reference, MLA relies on synthesizing narrative evidence spread across multiple reports.

\stitle{Comparison with CTIConnect pipelines.}
We further compare \system{} against CTIConnect's~\cite{cticonnect} vanilla RAG baseline and three specialized task-family pipelines using \code{gpt-5.4} on a stratified random sample of 150 main-set questions that preserves each task's original proportion. Evaluated under their released configurations over the identical corpus, these baselines comprise Extract-then-Retrieve for EL, Decompose-then-Retrieve for EA, and CSKG-guided RAG for MDS. As reported in Table~\ref{tab:pipelines}, \system{} trails the dedicated EL pipeline by just 0.003 while outperforming the EA pipeline by 0.025 and the MDS pipeline by 0.027, reaching an overall score of 0.836 compared with 0.823 across the combined task-specific pipelines and 0.705 for vanilla RAG. Rather than maintaining separate workflow architectures per task family, \system{} captures behavior decomposition, taxonomy restatement, and query-reformulation fallbacks as lightweight playbooks executed within a single unified agent across all nine tasks.

\begin{table}[htbp]
\papertablestyle
\caption{\rev{Mean scores for \system{} and CTIConnect pipelines
on 150 questions using \code{gpt-5.4}. Bold values indicate the highest score in each column.}}
\label{tab:pipelines}
\begin{tabularx}{\linewidth}{L c c c c}
\toprule
\textbf{System} & \textbf{EL} & \textbf{EA} & \textbf{MDS} & \textbf{Overall} \\
\midrule
Vanilla RAG & 0.699 & 0.527 & 0.831 & 0.705 \\
EtR / DtR / CSKG-guided & \best{0.926} & 0.610 & 0.828 & 0.823 \\
Base agent & 0.511 & 0.512 & 0.829 & 0.617 \\
\system{} & 0.923 & \best{0.635} & \best{0.855} & \best{0.836} \\
\bottomrule
\end{tabularx}
\end{table}

\stitle{Base-agent failures.}
\rev{To pinpoint where the base agent breaks down, we categorize incorrect answers from its recorded \code{gpt-5.4} trajectories on the main set into four failure stages. An error is labeled \emph{not located} when no tool output ever surfaces the record containing the target answer, \emph{located but not extracted} when the target record appears in a tool output yet the agent never identifies the gold identifier, \emph{extracted, wrong} when the agent surfaces the gold identifier during execution but omits it from the final answer, and \emph{Other} when the final output is unparseable.}

\begin{table}[t]
\papertablestyle
\setlength{\tabcolsep}{2.5pt}
\caption{\rev{Main-set failure distribution for the base agent
using \code{gpt-5.4}.}}
\label{tab:failclass}
\begin{tabularx}{\linewidth}{l c *{4}{C}}
\toprule
\textbf{Family} & \textbf{Failed} & \shortstack{\textbf{Not}\\\textbf{located}} & \shortstack{\textbf{Located,}\\\textbf{not}\\\textbf{extracted}} & \shortstack{\textbf{Extracted,}\\\textbf{wrong}} & \textbf{Other} \\
\midrule
EL & 212 & 0.24 & 0.47 & 0.22 & 0.07 \\
EA & 91 & 0.12 & 0.25 & 0.55 & 0.08 \\
\bottomrule
\end{tabularx}
\end{table}

\begin{table}[t]
\papertablestyle
\caption{\rev{Fraction of sampled base-agent entity-linking failures corrected, by class and configuration. Bold values indicate the highest recovery fraction in each row.}}
\label{tab:failfix}
\begin{tabularx}{\linewidth}{L c c c}
\toprule
\textbf{Base failure class} & Retrieval & \shortstack{\system{}\\w/o skill} & \system{} \\
\midrule
Not located & 0.18 & 0.55 & \best{0.82} \\
Located but not extracted & 0.09 & 0.82 & \best{0.95} \\
Extracted, wrong & 0.00 & 0.30 & \best{0.70} \\
\bottomrule
\end{tabularx}
\end{table}

In 47\% of EL failures, the base agent retrieves the relevant record yet fails to extract the gold identifier, whereas it fails to locate the record altogether in 24\%~(Table~\ref{tab:failclass}). On sampled failures where the record was located without extracting the identifier, retrieval alone resolves 9\%, typed tools without skills resolve 82\%, and the full system resolves 95\% (Table~\ref{tab:failfix}). By contrast, in 55\% of EA failures, the base agent surfaces the gold identifier mid-trajectory but omits it from the final response. The dominant bottleneck thus differs across task families, occurring during relation extraction from a retrieved record in EL versus candidate selection at answer submission in EA.

The remaining EL failure categories clarify the distinct contributions of typed tools and procedural skills. When the base agent fails to locate the target record, typed tools without skills recover 55\% of sampled failures compared with 18\% for retrieval alone, demonstrating the benefit of structured search and alias resolution. Conversely, when the base agent identifies the gold identifier during execution but omits it from the final answer, adding procedural skills to typed tools lifts recovery from 30\% to 70\% by enforcing answer-selection discipline. Because these recovery figures are conditioned on base-agent failures, the overall task scores in Table~\ref{tab:main} additionally reflect the questions that the base agent already answers correctly.

\subsection{RQ3: Tool Use, Cost, and Query Performance}
\label{sec:eval:efficiency}
\label{app:case}
\label{app:scale}

\begin{table*}[t]
\papertablestyle
\caption{Scale-set query profile for \system{} with \code{gpt-5.4}.
In (b), values in parentheses are relative shading scores from the source profile.
Dashes mark cells left blank in that profile.}
\label{tab:query-profile}
\captionsetup[subtable]{position=top,labelformat=parens,skip=2pt}
\begin{subtable}[c]{.32\linewidth}
  \centering
  \caption{Token use and cost by task family}
  \label{tab:query-cost}
  \begin{tabularx}{\linewidth}{@{}l *{3}{C}@{}}
  \toprule
  \textbf{Family} & \shortstack{\textbf{Input}\\\textbf{tokens/q}} &
    \shortstack{\textbf{Output}\\\textbf{tokens/q}} &
    \shortstack{\textbf{Total}\\\textbf{cost (\$)}} \\
  \midrule
  EL  & 16.3K & 270 & 15.22 \\
  EA  & 23.3K & 353 & 10.12 \\
  MDS & 10.4K & 741 &  4.88 \\
  \bottomrule
  \end{tabularx}
\end{subtable}\hfill
\begin{subtable}[c]{.66\linewidth}
  \centering
  \caption{Mean calls per question and first-call compliance}
  \label{tab:tool-profile}
  \setlength{\tabcolsep}{2pt}
  \begin{tabularx}{\linewidth}{@{}l *{9}{C}@{}}
  \toprule
   & \multicolumn{4}{c}{\textbf{EL}} & \multicolumn{2}{c}{\textbf{EA}} &
      \multicolumn{3}{c}{\textbf{MDS}} \\
  \cmidrule(lr){2-5}\cmidrule(lr){6-7}\cmidrule(lr){8-10}
  \textbf{Tool} & RCM & WIM & ATD & ESD & ATA & VCA & CSC & TAP & MLA \\
  \midrule
  \code{resolve\_entity}      & --  & --  & --  & --  & (20) & (11) & 1.1  & 1.4  & 1.3  \\
  \code{get\_entity}          & 1.4 & 2.1 & 1.1 & 1.2 & 2.7  & 3.5  & --   & (14) & (12) \\
  \code{ontology\_neighbors} & 1.1 & 1.9 & 1.0 & 1.1 & 1.1  & (9)  & --   & --   & --   \\
  \code{search\_kb}           & 1.1 & 2.9 & 1.1 & 1.0 & 2.0  & 2.3  & --   & --   & --   \\
  \code{search\_chunks}       & --  & --  & --  & --  & (13) & (11) & --   & --   & (10) \\
  \code{chunks\_mentioning}   & --  & --  & --  & --  & (9)  & --   & (14) & (15) & (25) \\
  \code{read\_report}         & --  & --  & --  & --  & (9)  & --   & 2.9  & 3.0  & 2.9  \\
  \midrule
  Skill compliance           & 1.00 & 1.00 & 1.00 & 1.00 & 0.45 & 0.90 & 1.00 & 1.00 & 1.00 \\
  \bottomrule
  \end{tabularx}
\end{subtable}
\end{table*}

\begin{table}[t]
\papertablestyle
\caption{Paired tool-call counts and overall scores on the main set.
Bold indicates the higher overall score in each model pair and in the
\code{gpt-5.4} ablation comparison.}
\label{tab:effort}
\setlength{\tabcolsep}{2.5pt}
\begin{tabularx}{\linewidth}{@{}L *{4}{c}@{}}
\toprule
 & \multicolumn{2}{c}{\textbf{Calls per question}} &
   \multicolumn{2}{c}{\textbf{Overall score}} \\
\cmidrule(lr){2-3}\cmidrule(lr){4-5}
\textbf{Model} & Base & \system{} & Base & \system{} \\
\midrule
\code{gpt-5.4}          & 3.76  & 3.77 & 0.609 & \best{0.832} \\
\code{gpt-5.4-mini}     & 5.30  & 5.23 & 0.578 & \best{0.754} \\
\code{sonnet-4-6}       & 9.82  & 4.59 & 0.564 & \best{0.769} \\
\code{haiku-4-5}        & 12.33 & 5.53 & 0.483 & \best{0.743} \\
\midrule
\textbf{\code{gpt-5.4} ablation} &
  \multicolumn{2}{c}{\textbf{Calls/q}} & \multicolumn{2}{c}{\textbf{Overall score}} \\
Tools only & \multicolumn{2}{c}{4.78} & \multicolumn{2}{c}{\best{0.720}} \\
Skill only & \multicolumn{2}{c}{3.95} & \multicolumn{2}{c}{0.674} \\
\bottomrule
\end{tabularx}
\end{table}

\begin{figure}[t]
\centering
\includegraphics[width=\linewidth]{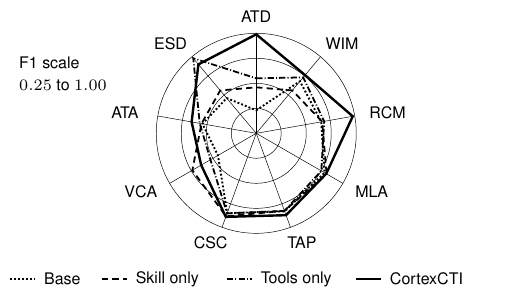}
\caption{Per-task scores for the \code{gpt-5.4} ablation on the main set.}
\label{fig:ablation-radar}
\end{figure}

\stitle{Tool use and investigation traces.}
Table~\ref{tab:effort} compares overall scores against mean tool invocations per question across the main set. On \code{gpt-5.4}, \system{} lifts the overall score by 0.223 while keeping mean invocations virtually unchanged from 3.76 to 3.77, and on \code{gpt-5.4-mini}, the score rises by roughly 0.18 as mean invocations edge down from 5.30 to 5.23. The Claude models exhibit substantial reductions in tool overhead, with \code{claude-haiku-4-5} gaining 0.260 while cutting mean invocations from 12.33 to 5.53, and \code{claude-sonnet-4-6} gaining 0.205 while cutting invocations from 9.82 to 4.59. Removing skills on \code{gpt-5.4} increases average invocations to 4.78 per question and lowers the score by 0.112 relative to the full system. Within both main configurations, lower-scoring runs correlate with longer trajectories, reflecting that harder questions induce additional exploratory calls.

Tool usage profiles on the scale set illustrate how \system{} adapts retrieval workflows across task families. For forward EL, the agent locates a source record via \code{search\_kb}, traverses its official edge via \code{ontology\_neighbors}, and inspects the target record via \code{get\_entity}, whereas for MDS it queries the entity index and reads roughly three full reports per question across CSC, TAP, and MLA. Table~\ref{tab:tool-profile} further quantifies how often the initial tool call matches the action prescribed by the task playbook, reaching 100\% on EL and MDS, 45\% on ATA, and 90\% on VCA.

\begin{figure*}[t]
\centering
\includegraphics[width=\linewidth]{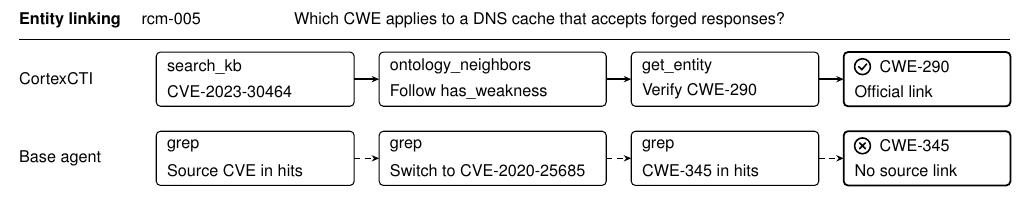}
\caption{Entity-linking case \code{rcm-005} using \code{gpt-5.4}.
\system{} follows the recorded weakness link from the shared source CVE.}
\label{fig:case}
\end{figure*}

Figure~\ref{fig:case} contrasts a EL case where both configurations issue the same number of calls yet reach opposite outcomes. On \code{rcm-005}, each agent executes three evidence-gathering calls before submission. Although the base agent locates the source CVE on its first \code{grep}, it subsequently drifts to an unrelated CVE and submits \code{CWE-345} after searching weakness files, whereas \system{} traverses the record's official cross-reference and verifies \code{CWE-290} before answering. Thus, while the base agent's lookup succeeds, subsequent unstructured searches sever its link to the originating record.

\stitle{Cost, latency, and accuracy on the scale set.}
On the scale set, we evaluate \system{} paired with \code{gpt-5.4} across 1{,}168 runs, representing roughly $1.7\times$ the main-set question count. Consistent with main-set behavior, each investigation averages roughly 2.6 US cents and 7 seconds, while the per-family costs in Table~\ref{tab:query-cost} total \$30.22 for query execution and \$32.08 when incorporating the one-time build, which amortizes to roughly 0.16 US cents per question across the 1{,}168 runs.

Input and output token consumption peak in distinct task families. Specifically, EA consumes the most input tokens at 23.3 thousand per question, compared with 16.3 thousand for EL and 10.4 thousand for MDS, whereas MDS generates the most output tokens at 741 per question, more than double the EA average of 353.

For accuracy, Table~\ref{tab:scale} reports results on the 1{,}076 held-out questions after excluding the 92 development items. Across these unseen questions, the unweighted nine-task average reaches 0.795 compared with 0.832 for the same configuration on the main set. EA experiences the steepest drop, falling from 0.646 to 0.540 across both attribution tasks, while mean EL F1 shifts from 0.908 to 0.869 as RCM climbs to 0.990 and WIM, ATD, and ESD decline by 0.033 to 0.067 relative to the main set. By contrast, mean MDS improves from 0.856 to 0.866, driven by gains on CSC and MLA that outweigh a slight decline on TAP.

\begin{table}[t]
\papertablestyle
\caption{\system{} scores on 1{,}076 held-out scale-set questions
using \code{gpt-5.4}, where $n$ is the number of questions per task.}
\label{tab:scale}
\begin{tabularx}{\linewidth}{L C C C}
\toprule
\textbf{Family} & \textbf{Task} & $n$ & \textbf{Score} \\
\midrule
\multirow{4}{*}{EL}
 & RCM & 175 & 0.990 \\
 & WIM & 192 & 0.695 \\
 & ATD & 148 & 0.957 \\
 & ESD & 166 & 0.833 \\
\midrule
\multirow{2}{*}{EA}
 & ATA & 92 & 0.588 \\
 & VCA & 119 & 0.492 \\
\midrule
\multirow{3}{*}{MDS}
 & CSC & 55 & 0.927 \\
 & TAP & 74 & 0.824 \\
 & MLA & 55 & 0.848 \\
\bottomrule
\end{tabularx}
\end{table}

\subsection{RQ4: Effects of Typed Tools and Skills}
\label{sec:eval:ablation}
\label{app:addbase:report}
\label{app:skilldecomp}
\label{app:addbase:retrieval}

\begin{table*}[t]
\papertablestyle
\caption{Main-set ablation of typed tools and skills using \code{gpt-5.4}.
Bold values indicate the highest score in each column. Underlining indicates the runner-up.}
\label{tab:ablation}
\begin{tabularx}{\linewidth}{l *{4}{C} *{2}{C} *{3}{C} C}
\toprule
 & \multicolumn{4}{c}{\textbf{EL}} & \multicolumn{2}{c}{\textbf{EA}} & \multicolumn{3}{c}{\textbf{MDS}} & \\
\cmidrule(lr){2-5}\cmidrule(lr){6-7}\cmidrule(lr){8-10}
\textbf{Config} & RCM & WIM & ATD & ESD & ATA & VCA & CSC & TAP & MLA & \textbf{Overall} \\
\midrule
\system{} & \best{0.980} & \best{0.760} & \best{0.990} & \second{0.900} & \best{0.656} & \second{0.636} & \best{0.889} & \best{0.871} & \second{0.808} & \best{0.832} \\
w/o skill                 & \second{0.690} & \second{0.730} & \second{0.550} & \best{0.980} & \second{0.574} & 0.529 & 0.852 & 0.825 & 0.751 & \second{0.720} \\
w/o tools                 & 0.670 & 0.560 & 0.460 & 0.560 & 0.556 & \best{0.738} & \second{0.883} & 0.824 & \best{0.812} & 0.674 \\
w/o skill \& tools        & 0.660 & 0.680 & 0.230 & 0.460 & 0.546 & 0.442 & 0.849 & \second{0.830} & 0.788 & 0.609 \\
\bottomrule
\end{tabularx}

\end{table*}

\stitle{Individual and combined effects.}
We vary two components in a $2{\times}2$ ablation while holding the harness, underlying model, and evaluation metrics fixed. Specifically, the tool condition replaces \code{bash} with our typed tools, whereas the skill condition injects the task-family playbook into the user turn. As shown in Table~\ref{tab:ablation}, skills alone lift the overall score from 0.609 to 0.674 and typed tools alone lift it to 0.720, whereas combining both reaches 0.832 for a net gain of 0.223, exceeding the sum of their individual gains by roughly 0.05.

Component contributions diverge across task families (Fig.~\ref{fig:ablation-radar}). Across the four EL tasks, mean F1 progresses from 0.508 for the base agent to 0.563 with skills alone, 0.738 with tools alone, and 0.908 with both combined. Removing typed tools causes sharp drops on ATD and ESD, which fall from 0.990 to 0.460 and from 0.900 to 0.560, respectively. Conversely, adding skills on top of typed tools elevates ATD from 0.550 to 0.990 and RCM from 0.690 to 0.980 while lowering ESD from 0.980 to 0.900, as trajectories lacking procedural playbooks occasionally query the target taxonomy directly and return spurious matches. On EA, the skills-only variant averages 0.647 compared with 0.646 for the full system, trading a lower ATA score for a higher VCA score, whereas MDS scores remain comparable across all four configurations.

In preliminary trials, smaller models adhered to the same skill instructions less reliably when placed in the system prompt rather than the user turn, so our ablation holds user-turn placement fixed across every configuration incorporating skills.

\stitle{Component-level ablations.}
To isolate the source of these gains, we decouple benchmark instructions from procedural guidance within the playbooks in Appendix~\ref{app:skills} and evaluate the base agent under two ablated prompt variants. Specifically, \emph{Base + benchmark adaptation} retains only task descriptions, answer formats, and output rules, whereas \emph{Base + procedural guidance} additionally supplies the investigation steps rewritten for execution via shell commands, with both variants evaluated using \code{gpt-5.4} on the stratified 150-question main-set sample.

As reported in Table~\ref{tab:skilldecomp}, benchmark adaptation alone improves both EA tasks while EL edges down from 0.511 to 0.498. Adding procedural guidance lifts EL to 0.699, whereas replacing shell commands with typed tools in the full system drives EL to 0.923, outperforming shell-based procedural guidance by 0.224. On EA, by contrast, the effect of typed tools varies across target taxonomies, as procedural guidance over the shell interface achieves 0.619 on ATA and 0.669 on VCA compared with 0.631 on ATA and 0.638 on VCA for the full configuration.

\begin{table}[htbp]
\papertablestyle
\caption{\rev{Scores with benchmark adaptation, procedural guidance, and
structured access on 150 sampled main-set questions. Bold values indicate the highest score in each column.}}
\label{tab:skilldecomp}
\begin{tabularx}{\linewidth}{L c c c c c}
\toprule
\textbf{Configuration} & \textbf{EL} & \textbf{ATA} & \textbf{VCA} & \textbf{MDS} & \textbf{Overall} \\
\midrule
Base & 0.511 & 0.510 & 0.514 & 0.829 & 0.617 \\
Base + benchmark adaptation & 0.498 & 0.557 & 0.549 & \best{0.857} & 0.630 \\
Base + procedural guidance & 0.699 & 0.619 & \best{0.669} & 0.818 & 0.726 \\
\system{} & \best{0.923} & \best{0.631} & 0.638 & 0.855 & \best{0.836} \\
\bottomrule
\end{tabularx}
\end{table}

These EL gains motivate isolating structured graph access by comparing retrieval with and without official-relation lookup. Evaluated without skills, the \emph{Retrieval} baseline equips the base agent with two search tools, namely \code{search\_kb} for hybrid dense and BM25 search over knowledge-base entries and \code{search\_chunks} for dense search over report passages, whereas \emph{Retrieval + relation lookup} adds \code{get\_entity} and \code{ontology\_neighbors} to traverse official cross-references. In Table~\ref{tab:retrieval}, unstructured retrieval alone raises EL F1 by just 0.027 over the base agent, whereas adding relation lookup yields a further 0.206 jump from 0.538 to 0.744. Because skills are omitted from both configurations, this contrast isolates the direct contribution of exposing official ontology edges through typed lookup tools.

\begin{table}[htbp]
\papertablestyle
\caption{\rev{Scores with retrieval and official-relation lookup
on 150 questions using \code{gpt-5.4}. Bold values indicate the highest score in each column.}}
\label{tab:retrieval}
\begin{tabularx}{\linewidth}{L c c c c}
\toprule
\textbf{Configuration} & \textbf{EL} & \textbf{EA} & \textbf{MDS} & \textbf{Overall} \\
\midrule
Base & 0.511 & 0.512 & 0.829 & 0.617 \\
Retrieval & 0.538 & 0.519 & 0.843 & 0.635 \\
Retrieval + relation lookup & 0.744 & 0.527 & 0.808 & 0.717 \\
\system{} w/o skill & 0.764 & 0.544 & 0.802 & 0.728 \\
\system{} & \best{0.923} & \best{0.635} & \best{0.855} & \best{0.836} \\
\bottomrule
\end{tabularx}
\end{table}

For MDS, we evaluate the joint contribution of alias matching and entity-indexed passage collection on the sampled questions. Specifically, the reduced configuration restricts \code{resolve\_entity} to canonical names and identifiers, removes \code{chunks\_mentioning}, and retains \code{read\_report}, \code{search\_chunks}, and the knowledge-base tools. As shown in Table~\ref{tab:reportabl}, the full Evidence Alignment Layer achieves 0.855 compared with 0.845 for the reduced variant. This 0.010 difference isolates the joint impact of cross-alias entity indexing when both configurations retain identical access to the underlying reports via chunk search and full-report reads. Whereas the earlier cluster evaluation assesses how accurately the index groups reports by adversary, this ablation directly measures how those groupings translate into claim accuracy in the final answer.

\begin{table}[htbp]
\papertablestyle
\caption{\rev{Evidence alignment ablation on sampled synthesis questions
using \code{gpt-5.4}. Bold values indicate the highest score in each column.}}
\label{tab:reportabl}
\begin{tabularx}{\linewidth}{L c}
\toprule
\textbf{MDS mean} & \textbf{\code{gpt-5.4}} \\
\midrule
Full alignment layer & \best{0.855} \\
No aliases or entity index & 0.845 \\
Difference & $+0.010$ \\
\bottomrule
\end{tabularx}
\end{table}

\subsection{RQ5: Transfer to Downstream Tasks}
\label{sec:eval:downstream}

\stitle{Vulnerability classification and technique attribution.}
To evaluate transfer to CTIBench~\cite{alam2024ctibench}, both configurations run mini-swe-agent with \code{gpt-5.4} under a 20-step budget over the identical reference records. While the shell baseline inspects flat files directly, \system{} operates with ontology access, lexical search, and matching procedural skills while keeping the report graph disabled. Across the first ten released examples from each task, mean identifier-set F1 climbs from 0.130 to 0.540 on CVE-to-CWE mapping and from 0.339 to 0.568 on ATT\&CK technique extraction, lifting the overall score from 0.235 to 0.554 for a gain of 31.9 points. A separate 16-item pilot on CTIConnect using the same model exhibits a comparable jump from 0.031 to 0.458.

\stitle{Security log triage and event localization.}
A standalone study compares BM25 top-five selections against up to five entries re-ranked by \code{gpt-5.4} from the same pool of at most 15 BM25 candidates under identical downstream models and prompts. On CERT r4.2~\cite{certinsider}, user triage over login and device logs raises positive-class F1 from 0.743 to 0.772 (paired 95\% CI$[-0.027, 0.085]$), , recovering 61 versus 55 of 70 positives while flagging 27 versus 23 of 140 controls as benign behavioral anomalies add false alarms. Across 64-record event-localization windows, Recall@5 shifts from 0.069 to 0.072 on BETH~\cite{highnam2021beth} ($n{=}59$) and from 0.096 to 0.086 on IoT-23~\cite{garcia2020iot23} ($n{=}78$), constrained by candidate-pool ceilings of 0.196 and 0.105 alongside stronger local event-rarity signals on BETH. On AIT-ADS~\cite{landauer2024aitads} ($n{=}66$), alert Recall@5 drops from 0.445 to 0.383 and nDCG@5 reaches 0.724 versus 0.844 for temporal ranking, reflecting window-based relevance criteria, while AuditBench~\cite{anand2026auditbench} ($n{=}25$) attack F1 drops from 0.533 to 0.400 (paired 95\% CI $[-28,+12]$ points) because sparse audit metadata obscures cross-event links. Overall, evidence re-ranking remains bounded by candidate-pool coverage and alignment between ranking criteria and task-specific relevance.

\stitle{Evidence-grounded answering and verification.}
The same ranking protocol evaluates passage retrieval for question answering on QASPER~\cite{dasigi2021qasper} and evidence retrieval for claim verification on SciFact~\cite{wadden2020scifact}. Across 60 completed pairs per benchmark, scorable gold evidence appears in 59 QASPER items under paragraph-level relevance per paper and in 40 SciFact items under abstract-level relevance across the corpus, with confidence intervals from 3{,}000 item-level paired bootstrap resamples. On QASPER, Recall@5 climbs from 0.409 to 0.603 (up 19.5 points, paired 95\% CI $[12.1, 27.1]$ points) while nDCG@5 rises from 0.343 to 0.609, and on SciFact, Recall@5 edges from 0.860 to 0.867 alongside an nDCG@5 gain from 0.790 to 0.845. By contrast, paired confidence intervals for downstream QASPER token F1 (0.305 to 0.356) and SciFact three-way label accuracy (0.750 to 0.733) span zero.

\section{Discussion}
\noindent\textbf{Frontier model scaling and capability tiers.}
We evaluate models across distinct capability tiers from two providers to test whether \system{} improves effectiveness without relying on a single frontier model. Holding the harness constant per model isolates the net gain of pairing structured CTI access with procedural skills at fixed capability. While newer flagship models may extract more evidence directly from flat files and narrow the relative margin, quantifying those gains remains future work.

\noindent\textbf{Temporal drift and corpus staleness.}
Investigations during emerging campaigns often face sparse reporting, while official taxonomy mappings may appear only after an inquiry begins. Although pinned snapshots and immutable report text keep records and character-level offsets reproducible, newly published reports and revised mappings remain inaccessible until ingestion. Therefore, empty retrieval results may reflect coverage lag rather than threat absence, whereas passages from superseded reports still provide valid historical evidence. Future extensions could index incoming and revised sources incrementally, expose version and coverage metadata to agents, and retain prior snapshots so existing citations remain auditable.

\noindent\textbf{Conflicting evidence across vendors.}
Multi-vendor CTI often exhibits analytical disagreement, with a seven-vendor study finding mostly moderate attribution consensus even after alias reconciliation~\cite{ethembabaoglu2026aptdisagree}. Vendors track campaigns at varying granularities, assign overlapping activity to distinct intrusion sets, or revise attributions with new telemetry, while separate reports may create false independent corroboration by echoing a shared upstream source. Although the evidence alignment layer excludes weak identity links from confirmed clusters and preserves source passages, entity merging resolves only identity, leaving claim compatibility and evidential independence to the agent. Future work could model claims with their provenance, temporal scope, and upstream dependencies to distinguish true contradiction, independent corroboration, and scope differences.

\section{Conclusion}
\label{sec:conclusion}
Existing threat intelligence workflows treat repositories as static stores and leave each agent to parse fragmented vendor reports and reference records on its own, struggling to trace claims to exact passages during multi-hop analysis. To address this, we presented CortexCTI, an agent-native knowledge infrastructure organized into a three-layer architecture that establishes a new paradigm for threat intelligence as a shared service. CortexCTI grounds canonical entities in validated ontology records, links cross-source mentions to character-verified spans, and exposes hybrid search, lookup, and graph traversal via typed tools and procedural skills. Across nine CTIConnect tasks and four models, it substantially boosts accuracy and cuts tool calls over flat-file access. Delivering this unified architecture as a service lets diverse harnesses reuse validated evidence and composable operations across investigations, advancing reliable, auditable threat analysis.
\subsection*{Ethical Considerations}
Using only public CTI records and CTIConnect summaries~\cite{cticonnect}, this study involves no human subjects, personal data, or proprietary victim information. Dual-use risks remain outweighed by defensive triage benefits since \system{} adds no non-public knowledge or offensive capability, while preserved source spans support evidence verification against misattribution. The authors declare no conflicts of interest.

\subsection*{Acknowledgment}
In this work, we used generative AI tools as an object of study and as a
component of the system. The experimental setup identifies the models
used for corpus construction and agent evaluation (\S\ref{sec:eval:setup}). We
additionally used generative AI assistance for writing support (copy-editing and
LaTeX formatting), and for coding support during implementation of the build
pipeline and evaluation scripts. We did not use generative AI tools to generate
research ideas, to produce experimental results, or to write or select the
related work. All AI-assisted code was reviewed and tested by the authors, and
all reported numbers come from executed runs. We have reviewed all AI-assisted
work and take responsibility for the final content of this paper, including its
text, claims, and artifacts.

\bibliographystyle{IEEEtran}
\bibliography{references}

\appendices
\section{Data Availability}
The CTIConnect experiments use its public benchmark and corpus~\cite{cticonnect}
and four public reference knowledge bases. Section~\ref{sec:eval:downstream}
cites the public datasets used in the downstream experiments. Source code for
the build pipeline, tool server, skills, and harness entry point is submitted
as anonymized supplementary material and will be released upon publication.

\section{Prompt Templates}
\label{app:prompts}

This appendix reproduces the agent system prompts, the build-time
operator prompts, and the MDS judge prompt verbatim. Placeholders in
\texttt{\{\}} (Python \texttt{str.format}) and \texttt{\{\{\ \}\}} (Jinja2) are
substituted at runtime. Throughout,
\colorbox{pastelbluebg}{\strut\small system prompts} are shown in blue,
\colorbox{pastelgreenbg}{\strut\small user prompts} in green, and
\colorbox{pastelrosebg}{\strut\small skill playbooks} in rose.

\subsection{Agent System Prompts: the Two Arms}
\label{app:prompt_agent}

Both arms run
the same harness, model, step budget, and temperature; the only difference
between them is which of the following is installed as the system prompt and
which action surface it describes---the seven typed tools of
Table~\ref{tab:tools} for \system{}, a single \code{bash} tool over the corpus
dumped to flat files for the base arm. Each prompt is deliberately short: the
substrate, not the prompt, is what the paper varies.
\begin{systempromptbox}[System Prompt --- \system{} Arm (typed tools)]
\begin{lstlisting}[style=promptstyle]
You are a CTI analyst. Investigate the question using the available tools, one or more tool calls per turn, then call submit_answer with the final answer. Resolve names to canonical entities first; prefer authoritative structure over inference; state exact identifiers (CVE-/CWE-/CAPEC-/T-ids, canonical names) explicitly.
\end{lstlisting}
\end{systempromptbox}
\begin{systempromptbox}[System Prompt --- Base Arm (bash over flat files)]
\begin{lstlisting}[style=promptstyle]
You are a CTI analyst with a bash shell. The current directory holds a flat CTI corpus: reports/<doc_id>.txt (321 vendor threat reports) and kb/{cve,cwe,capec,technique,tactic}.jsonl (one JSON entry per line). Investigate with one bash command per turn (grep/cat/awk; each command runs in a fresh shell, so chain with pipes). State exact identifiers (CVE-/CWE-/CAPEC-/T-ids, canonical names) explicitly. When you have the answer, run a single command whose FIRST output line is exactly COMPLETE_TASK_AND_SUBMIT_FINAL_OUTPUT followed by your answer text, e.g.
  echo COMPLETE_TASK_AND_SUBMIT_FINAL_OUTPUT; echo 'The weakness is CWE-384.'
\end{lstlisting}
\end{systempromptbox}

The user turn is identically templated in both arms; the per-task skill of
Appendix~\ref{app:skills} is prepended to it for the \system{} arm, and the
answer-format template for the MDS tasks is appended in both arms so that the
two arms are scored on the same output shape.
\begin{userpromptbox}[User Turn Template --- Both Arms]
\begin{lstlisting}[style=promptstyle]
## Question
{{task}}
\end{lstlisting}
\end{userpromptbox}

\subsection{Build Time: Span-Grounded Extraction}
\label{app:prompt_extract}

The report is presented as numbered spans and every emitted surface must be a verbatim
substring of the span it is attributed to, which is what makes the
character-offset provenance checkable: the
build validator re-locates each surface in its span and drops what it cannot
find.
\begin{systempromptbox}[System Prompt --- Span-Grounded Extraction]
\begin{lstlisting}[style=promptstyle]
You are a precise CTI information-extraction engine. Output only JSON.
\end{lstlisting}
\end{systempromptbox}
\begin{userpromptbox}[User Prompt --- Span-Grounded Extraction]
\begin{lstlisting}[style=promptstyle]
Extract a span-grounded knowledge graph from this cyber threat intelligence report.

The report is segmented into numbered spans:
{spans_block}

Emit a JSON object:
{{
  "mentions": [
    {{"id": "m1", "entity_type": "<one of {etypes}>",
      "surface": "<verbatim substring of the span>", "sent_idx": <span number>}}
  ],
  "triples": [
    {{"src": "m1", "rel": "<one of {rels}>", "dst": "m2", "sent_idx": <span number where this relation is asserted>}}
  ]
}}

Rules:
- "surface" MUST be copied verbatim (case included) from the span numbered sent_idx.
- COMPLETENESS MATTERS: extract EVERY named threat actor, malware family, attack tool, technique, vulnerability, campaign, indicator (hash/IP/domain), targeted sector/region/organization — including ones mentioned in passing (e.g. "recruited affiliates from BlackMatter, REvil and DarkSide" names three threat actors).
- A triple's relation must be explicitly asserted in its span, not inferred from co-occurrence.
- "alias_of" triples capture EVERY naming statement: "X (also known as Y)", "X, tracked as Y", "X aka Y", "the X group operates the Y ransomware" does NOT make X an alias of Y, but "Y (formerly X)" does. Never miss an alias statement — cross-vendor naming is critical downstream.
- Entities, not descriptions: surfaces should be proper names or identifiers, not generic phrases ("the malware", "a phishing campaign" are NOT mentions; ransom amounts are NOT indicators).
- Prefer specific relations; emit nothing you cannot anchor to a span.

Worked example. Span "[2] DarkGate (also known as MehCrypter), operated by the Rastafareye persona, exploited CVE-2024-21412 to target financial organizations in Europe" yields mentions m1=DarkGate/malware, m2=MehCrypter/malware, m3=Rastafareye/threat_actor, m4=CVE-2024-21412/vulnerability, m5=financial/sector, m6=Europe/region and triples (m2 alias_of m1), (m3 operates m1), (m1 exploits m4), (m1 targets m5), (m1 targets m6), all with sent_idx=2.
\end{lstlisting}
\end{userpromptbox}

\subsection{Build Time: TTP Extraction}
\label{app:prompt_ttp}

The grounding operator runs without a system prompt.
The rules encode the two error modes
that dominate uncurated ATT\&CK tagging---defensive recommendations read as
attacker behavior, and outcome-phrased behaviors missed entirely---and the
no-invention rule feeds the zero-fabrication invariant, since every emitted id
is checked against the materialized ATT\&CK node set at validation time.
\begin{userpromptbox}[User Prompt --- TTP Extraction]
\begin{lstlisting}[style=promptstyle]
Extract MITRE ATT&CK techniques from this CTI report chunk.

Chunk:
"""{chunk}"""

The 14 ATT&CK tactics (for scope): {tactics}

List every ATTACKER behavior in the chunk that corresponds to a real ATT&CK
technique. For each: "behavior" = the verbatim phrase; "technique_id" = the
ATT&CK id (e.g. T1486 or T1059.001), most specific you are confident in;
"technique_name" = official name.

STRICT rules:
- Attacker techniques ONLY. EXCLUDE defenses/mitigations/recommendations
  ("enforce MFA", "monitor logs", "apply patches", "network segmentation",
  "third-party risk management"), generic outcomes ("data theft", "extortion"),
  and bare tool/malware names.
- Include techniques stated concisely as OUTCOMES. Examples:
  "encrypts files / systems" -> T1486 Data Encrypted for Impact;
  "deletes volume shadow copies / inhibits recovery" -> T1490;
  "disables/kills security tools" -> T1562.001 Disable or Modify Tools;
  "uses valid/compromised/stolen accounts" -> T1078 Valid Accounts;
  "exploits a public-facing app / VPN / web vulnerability" -> T1190;
  "spearphishing / phishing email" -> T1566;
  "PowerShell / command-line execution" -> T1059.
- Do NOT invent ids. If unsure of the exact id, omit the behavior.

Return JSON: {{"ttps": [{{"behavior": "...", "technique_id": "T....", "technique_name": "..."}}]}}.
\end{lstlisting}
\end{userpromptbox}

\subsection{Build Time: Entity-Resolution Adjudication}
\label{app:prompt_resolve}

Deterministic signals resolve most mentions; this judge adjudicates only the
residue. It is the operator behind the canonical
cross-vendor entities, and its strictness is the safeguard against the
mega-cluster failure discussed in \S\ref{sec:eval:build}: near-miss names
(BlackCat vs.\ BlackMatter) and shared-sponsor actors (Lazarus, Kimsuky,
Andariel) must stay distinct, and a span that merely lists both names is
treated as evidence of difference rather than sameness. The three-valued
\texttt{kind} field lets downstream merging accept only the authoritative tier.
\begin{systempromptbox}[System Prompt --- Entity-Resolution Judge]
\begin{lstlisting}[style=promptstyle]
You are a CTI entity-resolution judge. Decide whether two entity mentions from different threat reports refer to the same real-world entity. Output only JSON.
\end{lstlisting}
\end{systempromptbox}
\begin{userpromptbox}[User Prompt --- Entity-Resolution Judge]
\begin{lstlisting}[style=promptstyle]
Mention A: "{sa}" (type: {ta}, vendor: {va})
  supporting span: "{ctx_a}"

Mention B: "{sb}" (type: {tb}, vendor: {vb})
  supporting span: "{ctx_b}"

Are A and B the same real-world entity? Cross-vendor naming differs (e.g. APT29 = Cozy Bear = Midnight Blizzard), but distinct entities often have similar names (e.g. BlackCat vs BlackMatter are DIFFERENT ransomware families).

Strict rules:
- A span that merely LISTS both names among several actors, or says one "overlaps with" / "may be confused with" / "is distinct from" the other, asserts they are DIFFERENT entities, not the same.
- Answer "exact" only when the names are well-established canonical aliases of one entity, or a span contains an explicit naming statement ("aka", "also known as", "tracked as", "formerly") directly connecting A and B.
- Sharing a country, sponsor, toolset, or campaign does NOT make two actors the same (Lazarus, Kimsuky and Andariel are all DPRK groups and all DIFFERENT).

JSON: {{"same": true/false, "kind": "exact" | "claimed-same" | "possibly-same", "reason": "<one sentence>"}}
- "exact": an authoritative naming relationship is evident
- "claimed-same": a span itself asserts the equivalence
- "possibly-same": contextual evidence warrants association but not certainty
\end{lstlisting}
\end{userpromptbox}

\subsection{Build Time: Triple Validation}
\label{app:prompt_triple}

This prompt belongs to an earlier stage of development that materialized report
triples. It is not part of the evaluated build.

The verifier scores each candidate report-layer edge against the single
sentence offered as evidence. Restricting admissible evidence to
that one span is what separates an asserted relation from a co-occurrence, and
the type-correction fields let the judge repair an entity typing without
discarding the edge.
\begin{systempromptbox}[System Prompt --- Triple-Validation Judge]
\begin{lstlisting}[style=promptstyle]
You are a strict CTI fact-verification judge. Score whether a single sentence supports a single relational claim. Output only JSON.
\end{lstlisting}
\end{systempromptbox}
\begin{userpromptbox}[User Prompt --- Triple-Validation Judge]
\begin{lstlisting}[style=promptstyle]
Claim: ({src} [{src_type}]) --{rel}--> ({dst} [{dst_type}])

The ONLY admissible evidence is this sentence from a {vendor} report:
"{span_text}"

Score on three criteria:
1. predicate explicitness — is the relation "{rel}" explicitly asserted in the sentence, or merely inferred from co-occurrence?
2. entity scope — are both entities within the predicate's syntactic scope in this sentence?
3. CTI semantic validity — is "{rel}" type-compatible with a {src_type} and a {dst_type}? Are the entity types themselves correct (e.g. REvil is a threat actor AND a ransomware; "recruits affiliates from X" does not mean "uses X")?

JSON: {{"score": <0.0-1.0>, "reason": "<one sentence>",
       "src_type_correction": null | "<corrected type>",
       "dst_type_correction": null | "<corrected type>"}}
\end{lstlisting}
\end{userpromptbox}

\subsection{Evaluation: Multi-Document Synthesis Judge}
\label{app:prompt_judge}

The MDS family (CSC, TAP, MLA) is free-form, so it is scored by the
claim-coverage judge below rather than by identifier F1. Matching is
many-to-many and semantically tolerant:
the judge decomposes both answers into atomic claims, then judges each side
independently, which is what makes the metric robust to a prediction that
splits or merges the reference's sentences. It is applied identically to both
arms, and \S\ref{sec:eval:build} reports its measured resolution ($0.06$), the
threshold below which we decline to read an MDS gap as real.
\begin{userpromptbox}[Judge Prompt --- MDS Claim Coverage]
\begin{lstlisting}[style=promptstyle]
You are an expert Cyber Threat Intelligence (CTI) analyst acting as an
impartial judge. You will score a model's free-form answer to a
multi-document synthesis question against a reference (gold) answer, using
claim-level COVERAGE matching (v2: many-to-many, semantically tolerant).

## Method

1. Decompose the REFERENCE answer into a list of atomic claims. An atomic
   claim is a single, self-contained, verifiable statement (one fact about a
   threat actor name/alias, one TTP, one target, one tool, one date, one
   capability, etc.). Do not merge multiple facts into one claim.

2. Decompose the PREDICTION answer into atomic claims using the same rule.
   IMPORTANT: a compound prediction sentence containing several facts MUST be
   split into one claim per fact (same granularity as the reference side).

3. Coverage matching — NOT one-to-one. Judge each side independently:
   - A REFERENCE claim is COVERED if its content is expressed by ANY
     prediction claim, or jointly by SEVERAL prediction claims.
   - A PREDICTION claim is SUPPORTED if its content is expressed by ANY
     reference claim, or is part of the content of ONE reference claim.
     When two or more prediction claims together correspond to one reference
     claim, ALL of them count as supported.

4. Semantic tolerance — the following count as a MATCH:
   - paraphrase and alias equivalence (e.g. "APT29" matches "Cozy Bear";
     "spear-phishing" matches "targeted phishing emails");
   - temporal granularity and qualifier differences when the core fact
     (year, entity, event) agrees: "since 2021" ≈ "late 2021",
     "early 2024" ≈ "January 2024", "April 2025" ≈ "April 23, 2025";
   - singular/plural, word order, and one-word qualifier differences that do
     not change the identified entity, event, or time.
   Do NOT match claims about different entities, different events, or
   clearly different facts.

## Question
{{ QUESTION }}

## Reference (gold) answer
{{ REFERENCE }}

## Prediction (model) answer
{{ PREDICTION }}

## Output

Return ONLY a JSON object, no prose, no code fences:

{
  "reference_claims": ["<atomic claim>", ...],
  "prediction_claims": ["<atomic claim>", ...],
  "covered_reference_indices": [<0-based indices of covered reference claims>],
  "supported_prediction_indices": [<0-based indices of supported prediction claims>]
}
\end{lstlisting}
\end{userpromptbox}

\newpage

\section{Procedural Skill Playbooks}
\label{app:skills}

The three procedural skills ship as five Markdown
playbooks: entity linking and attribution each carry a specialization for the
subtask whose direction or output type differs from its family default, and
synthesis serves all three MDS subtasks. The router is a static
task-to-playbook map---\code{rcm}, \code{atd}, \code{esd} $\rightarrow$ entity
linking; \code{wim} $\rightarrow$ entity linking (reverse); \code{ata},
\code{vca} $\rightarrow$ the two attribution playbooks; \code{csc},
\code{tap}, \code{mla} $\rightarrow$ synthesis---with no model call and no
per-question adaptation. Each playbook is injected verbatim into the user turn
rather than the system prompt. Smaller models followed the same text less
reliably in the system prompt (\S\ref{sec:eval:ablation}).

The ablation removes the playbooks in the
\emph{w/o skill} arm; the \emph{w/o tools} arm keeps them but serves them over
bash, which is why their central prescriptions---resolve, traverse an official
edge, consult an alias set---become inert there.

\rev{\stitle{Benchmark adaptation lines.} Besides procedure, each playbook contains lines that
adapt the agent to CTIConnect: how the task is posed and answered, and its rules.
Entity linking: ``The description paraphrases one specific SOURCE entry'',
``NEVER answer empty'', ``Answer with exactly one identifier''.
Reverse entity linking: ``The answer is always a CVE identifier'',
``Answer with exactly one CVE identifier'', ``NEVER answer empty''.
Attribution to ATT\&CK: ``The answer is exactly one T-id'',
``a wrong extra identifier costs exactly as much as a miss'',
``your final answer is scored as one of your predictions''.
Attribution to CWE: ``The answer is always a CWE'',
``abstaining and answering wrongly score the same'',
``a wrong extra identifier costs exactly as much as a miss''.
Synthesis: the note that unsupported claims lower the score.}

\subsection{Entity Linking (RCM, ATD, ESD)}
\label{app:skill_el}
\begin{skillpromptbox}[Skill Playbook --- \texttt{entity-linking.md}]
\begin{lstlisting}[style=promptstyle]
# Skill: cross-taxonomy entity linking

Trigger: a behavioral description must be mapped to an entry in another CTI
taxonomy. The source entry is paraphrased but its ID is hidden.

**The single most important rule: do NOT search the TARGET taxonomy first.**
The description paraphrases one specific SOURCE entry; the authoritative
cross-reference edge from that source entry gives the answer. Searching the
target taxonomy directly falls into the cross-source vocabulary gap and picks
plausible-but-wrong entries.

Routing table — identify the question form, then execute:

| Question form | Step 1 (mandatory first call) | Step 2 |
|---|---|---|
| vulnerability description → "which CWE" | `search_kb(query=<description>, kb="cve")` | `ontology_neighbors(node_id=<CVE>, rel="has_weakness")` |
| weakness description → "which CVE instantiates" | keyword-probe loop over `search_kb(kb="cve", must_terms=[...])` — see below | confirm the winner's CWE via `ontology_neighbors(node_id=<CVE>, rel="has_weakness")` matches the described weakness |
| attack-pattern description → "which ATT&CK technique" | `search_kb(query=<description>, kb="capec")` | `ontology_neighbors(node_id=<CAPEC>, rel="maps_to_technique")` |
| weakness description → "which CAPEC exploits it" | `search_kb(query=<description>, kb="cwe")` | `ontology_neighbors(node_id=<CWE>, rel="exploits_weakness", direction="in")` |

Keyword-probe loop (weakness→CVE): the question paraphrases the target CVE's
description, so they share rare discriminative vocabulary. Run TWO probes and
cross-check — never trust a single retrieval path:
(1) Path A: `search_kb(kb="cve", query=<question>)` with NO must_terms — note
    the top-5 (semantic + lexical ranking).
(2) Path B: pick the 2-3 most specific technical terms in the question
    (component names, mechanism words like "hardware random", "sandbox",
    instruction/protocol names — never generic words like system/attacker),
    then `search_kb(kb="cve", query=<question>, must_terms=[t1, t2])`.
    Iterate on `n_term_matches_in_kb`: >30 hits → add a term; 0 hits → drop
    the weakest term or swap a synonym ("randomness"→"entropy"); give up
    after 3 probe rounds and fall back to Path A's list.
(3) Path C: if your own knowledge suggests a specific CVE id for this
    description, verify it — `get_entity` on that id and check its
    description against the question. Memory is a candidate generator, never
    an answer by itself.
    Candidates appearing in MULTIPLE paths are strongest. A unique single-path
    hit is a good LEAD, not an answer — it still must pass step (4).
(4) FINAL CHECK (mandatory, per candidate): read the description and tick off
    EVERY specific detail of the question — mechanism, component, attack
    consequence. Pick the candidate matching ALL details; if none matches
    all, pick the one matching the mechanism (not the component). Recency or
    CVSS is NOT a tiebreaker. Each question is independent — never reuse the
    previous question's CVE just because the wording feels similar.
(5) Confirm via the CVE's `has_weakness` CWE edge when present (~9% of CVEs have no edge — a missing edge is NOT disconfirmation); NEVER answer empty — if
    unresolved after all probes, answer Path A's best mechanism match.

Step 3 — verify before answering: `get_entity` on the candidate target; its
title/description must actually match the question's behavior. If the top
source candidate's neighbors contain no plausible target, try the next source
candidate from step 1 (the right source is usually in the top 3).

Fallback only when step 2 returns nothing for all plausible sources:
`search_kb` over the TARGET taxonomy with keyphrases restated in that
framework's idiom (CWE: "Improper/Missing/Incorrect X"; ATT&CK:
"Parent: Sub-technique" noun phrases; CAPEC: attack-method verb phrases),
then verify with `get_entity`.

Answer with exactly one identifier, stated explicitly, plus one sentence of
reasoning grounded in the verified entry.
\end{lstlisting}
\end{skillpromptbox}

\subsection{Entity Linking, Reverse Direction (WIM)}
\label{app:skill_wim}

WIM inverts the linking direction (weakness description $\rightarrow$
instantiating CVE) and is the one EL subtask on which the scaffold does not
approach ceiling. It gets its own playbook because the family rule---search the
source taxonomy---inverts with it: here the source is the CWE, and the
answer set is the reverse \code{has\_weakness} edge.
\begin{skillpromptbox}[Skill Playbook --- \texttt{entity-linking-wim.md}]
\begin{lstlisting}[style=promptstyle]
# Skill: weakness description → the CVE that instantiates it

Trigger: the question paraphrases one CWE's definition and asks which CVE
instantiates that weakness. The answer is always a CVE identifier.

**The single most important rule: do NOT search the CVE corpus first.**
The question paraphrases a CWE, not a CVE. CVE descriptions name products and
mechanisms, never the weakness class, so lexical overlap between the question
and the right CVE is weak and misleading. Go through the CWE.

Step 1 — locate the SOURCE CWE: `search_kb(query=<the description>, kb="cwe")`.
The description is a close paraphrase of one CWE entry, so the right entry
normally lands in the top 3. Confirm with `get_entity` that its definition
covers every element of the description before moving on.

Step 2 — walk the authoritative edge:
`ontology_neighbors(node_id=<CWE>, rel="has_weakness", direction="in")`.
This enumerates exactly the CVEs that NVD records as instances of that
weakness. It is the answer set, not a hint: every member is a defensible
answer to "which CVE instantiates this weakness". Do not discard it and go
searching the CVE corpus instead — that trades an authoritative answer for a
guess.

Step 3 — pick one from that set. Every member already satisfies the weakness,
so read the candidates' descriptions with `get_entity` and prefer the one that
also matches the question's extra specifics (affected component, attack
consequence, privilege required). With nothing to separate them, answer the
first. Recency and CVSS are NOT tiebreakers.

If step 1 yields no convincing CWE, try the next CWE candidate from the search
(the right source is usually in the top 3) before giving up on this route.

Fallback — ONLY when no plausible CWE exists or its reverse edge is empty
(some CWEs have no linked CVE). Search the CVE corpus directly:
(1) Path A: `search_kb(kb="cve", query=<question>)` with NO must_terms — note
    the top-5.
(2) Path B: pick the 2-3 most specific technical terms in the question
    (component names, mechanism words like "hardware random", "sandbox",
    instruction/protocol names — never generic words like system/attacker),
    then `search_kb(kb="cve", query=<question>, must_terms=[t1, t2])`.
    Iterate on `n_term_matches_in_kb`: >30 hits → add a term; 0 hits → drop
    the weakest term or swap a synonym ("randomness"→"entropy"); give up
    after 3 probe rounds and fall back to Path A's list.
(3) Read each candidate's description and tick off EVERY specific detail of
    the question — mechanism, component, consequence. Pick the candidate
    matching ALL details; if none matches all, pick the one matching the
    mechanism (not the component).

Answer with exactly one CVE identifier, stated explicitly, plus one sentence
of reasoning grounded in the verified entry. The answer to this question is a
CVE: a CWE or CAPEC id is an intermediate hop, never the answer — if the id
you are about to submit does not start with `CVE-`, you stopped early, so go
back to Step 2 and continue. NEVER answer empty.
\end{lstlisting}
\end{skillpromptbox}

\subsection{Attribution to ATT\&CK Techniques (ATA)}
\label{app:skill_ata}

ATA and VCA share a decompose--restate--verify spine but differ in output
cardinality and target taxonomy, so they ship separately. Both encode the
scoring geometry explicitly: under symmetric identifier F1 a spurious
identifier costs exactly what a miss does, hence the rule that rejected
candidates must not appear anywhere in the answer text.
\begin{skillpromptbox}[Skill Playbook --- \texttt{attribution-ata.md}]
\begin{lstlisting}[style=promptstyle]
# Skill: grounding a report narrative to its ATT&CK technique

Trigger: a quoted blog passage describes attacker behavior; identify the
ATT&CK technique it maps to. **The answer is exactly one T-id.**

Every question here has a single gold technique. A second plausible T-id
cannot gain you anything and strictly costs precision, so pick the best one
and drop the runner-up -- even when two feel equally good. A passage often
narrates several steps (delivery, execution, evasion); the question asks for
the technique it is *about*, the one the passage spends its detail on, not
every step you can name.

Recommended sequence:
1. **Decompose the passage into atomic behaviors** (each a single actionable
   security event). Prose interleaves several behaviors; one-to-many is common.
2. Per behavior, `search_kb(kb="mitre", top_k=10)` — the right entry often ranks 5-10, NOT top-3, because
   famous sibling techniques outrank precise ones. Restate the behavior in
   the taxonomy's own idiom — ATT&CK names are "Parent: Sub-technique" noun
   phrases ("Subvert Trust Controls: Mark-of-the-Web Bypass"); CWE names are
   "Improper/Missing/Incorrect X". Narrative verbs rarely match: canonicalize
   before searching.
   When you pick a parent technique, also call
   `ontology_neighbors(node_id=<T-id>, rel="sub_technique_of",
   direction="in")` and check whether a sub-technique matches the passage's
   specifics better.
3. **Validate each candidate** with `get_entity`: the entry's description must
   cover the described behavior, not merely share words. Reject co-occurrence
   matches. Sub-technique beats parent technique when the detail supports it.
4. If the passage names identifiers (CVE-...), `ontology_neighbors` from them
   (has_weakness) gives authoritative CWE anchors for free.
5. Commit to ONE technique. If two candidates survive verification, choose on
   the passage's own wording rather than listing both. If none convincingly
   matches, re-search with 2-3 alternative phrasings before settling.
   Answer any sub-questions (platforms, data sources) from the `get_entity`
   attrs — they are authoritative fields, not guesses.

Decision points:
- Behavior matches both parent and sub-technique → prefer the sub-technique if
  the passage's specifics warrant it, else the parent.
- Precision over recall: a wrong extra identifier costs exactly as much as a
  miss. Never present "related" or "additionally relevant" identifiers.
- CRITICAL OUTPUT RULE: every CWE-/T- identifier token appearing ANYWHERE in
  your final answer is scored as one of your predictions. Mention ONLY your
  chosen identifier; never name rejected candidates, comparisons, or
  alternatives in the answer text. Exactly one T-id may appear anywhere in
  the answer.
\end{lstlisting}
\end{skillpromptbox}

\subsection{Attribution to CWE Weaknesses (VCA)}
\label{app:skill_vca}
\begin{skillpromptbox}[Skill Playbook --- \texttt{attribution-vca.md}]
\begin{lstlisting}[style=promptstyle]
# Skill: grounding a vulnerability narrative to its CWE

Trigger: a quoted blog passage describes a vulnerability or exploitation
narrative; identify the CWE weakness it maps to. **The answer is always a CWE.**

A passage about attacker behavior will always suggest plausible ATT&CK
techniques too — ignore them. Search `kb="cwe"` only, and emit only `CWE-nnn`
identifiers. A T-id in the answer is not a partial credit, it is a wrong
prediction. If the passage feels too thin to ground a CWE confidently, still
commit to the best-supported one: abstaining and answering wrongly score the
same, so a calibrated answer strictly dominates.

If the question truncates mid-sentence (e.g. ends at "Please provide: 1)"),
answer with the single CWE identifier plus one sentence of justification.

Recommended sequence:
1. **Decompose the passage into atomic behaviors** (each a single actionable
   security event). Prose interleaves several behaviors; one-to-many is common.
2. Per behavior, `search_kb(kb="cwe", top_k=10)` — the right entry often
   ranks 5-10, NOT top-3, because famous sibling weaknesses outrank precise
   ones. Restate the behavior in CWE's own idiom: entries are named
   "Improper/Missing/Incorrect X". Narrative verbs rarely match: canonicalize
   before searching.
3. **Validate each candidate** with `get_entity`: the entry's description must
   cover the described behavior, not merely share words. Reject co-occurrence
   matches.
4. If the passage names identifiers (CVE-...), `ontology_neighbors` from them
   (has_weakness) gives authoritative CWE anchors for free.
5. Consolidate to a CALIBRATED set: output the smallest set of identifiers
   that covers every described behavior. One behavior -> usually one identifier;
   but when two candidates BOTH plausibly match a behavior after verification
   and you cannot separate them on the entry text, include both. Do not include
   a third. If NO candidate convincingly matches a behavior, re-search with
   2-3 alternative phrasings (different idiom, different aspect of the
   behavior) before settling.
   Answer any sub-questions (platforms, data sources) from the `get_entity`
   attrs — they are authoritative fields, not guesses.

Decision points:
- The gold answer is usually the weakness CLASS the passage illustrates, not
  the narrowest variant you can find. When a specific CWE and its more general
  parent both fit, prefer the one the passage's own wording supports; do not
  reach for a narrower variant on detail the passage never states.
- Precision over recall: a wrong extra identifier costs exactly as much as a
  miss. Never present "related" or "additionally relevant" identifiers.
- CRITICAL OUTPUT RULE: every CWE-/T- identifier token appearing ANYWHERE in
  your final answer is scored as one of your predictions. Mention ONLY your
  chosen identifier(s); never name rejected candidates, comparisons, or
  alternatives in the answer text. Before submitting, delete every T-id from
  the answer text — including ones cited only as supporting context.
\end{lstlisting}
\end{skillpromptbox}

\subsection{Multi-Document Synthesis (CSC, TAP, MLA)}
\label{app:skill_mds}

One playbook serves all three MDS subtasks. It is the only skill whose
prescriptions are mostly about \emph{coverage} rather than routing, which
matches the per-tool profile of Table~\ref{tab:tool-profile}: the MDS agent leans
on \code{read\_report} and the entity index and never touches the ontology
tools.
\begin{skillpromptbox}[Skill Playbook --- \texttt{synthesis.md}]
\begin{lstlisting}[style=promptstyle]
# Skill: multi-report synthesis (actor profiles, malware lineage, campaign timelines)

Trigger: a question about one threat entity whose intelligence is scattered
across several vendor reports — possibly under different names.

Recommended sequence:
1. **Cover the listed cluster first**: if the question lists a report
   cluster (ids like [BLOG-n]), fetch EVERY listed report up front —
   `read_report` on each id (bash arm: cat reports/<id>.txt). These reports
   ARE the question's scope; do not skip any of them.
2. **Resolve the anchor entity**: `resolve_entity` on the entity named in the
   question (or surface it via `search_chunks` first if only behavior is
   given). The result lists the build-aggregated cross-vendor `aliases` and
   `n_docs` — for alias/naming questions, that aliases field is the complete
   authoritative set.
3. **Collect additional content through the entity index**:
   `chunks_mentioning` on the resolved entity id lists every report where
   any alias appears (doc_id, vendor, excerpt) — `read_report` any listed
   report not yet read. Also `resolve_entity` on related names found along
   the way (variants, predecessor families). Stay on the campaign the
   question is about: drop reports that merely share an actor name but
   describe a different campaign.
4. **Per-report checklist extraction** (do NOT summarize the pile in one
   pass): first list the distinct reports you retrieved; then for EACH
   report, go through EVERY field of the required answer format and note what
   that report contributes (names, dates, capabilities, targets).
   `read_report` starts with the report's build-extracted
   `indexed_entities`/`indexed_ttps` — a handy floor for what the report
   mentions (type buckets are heuristic) — then read the full text: facts
   often sit in passages that do not name the anchor entity.
5. **Merge across reports/vendors**: union the per-report notes field by
   field; group by vendor for corroboration. For dates and timelines, use
   only dates stated in the report TEXT (activity dates, disclosure dates in
   prose) — metadata is not provided. Copy date qualifiers verbatim
   ("late 2021", "early 2024", "April 23, 2025") — do not round or reword.
6. **Synthesize with explicit alias resolution**: name the canonical entity,
   list the aliases and which vendor uses which, order events by the dates
   stated in the text, attribute claims to vendors. Note disagreements
   rather than averaging them.
7. **Answer form**: compact bullets answering EXACTLY the aspects asked
   (e.g. "canonical name / aliases resolved" or "dates / phases"). Every
   bullet must be a claim grounded in a retrieved chunk. No background
   filler, no speculation, no "additionally" padding — extra unsupported
   claims directly lower your score.

Decision points:
- "All / every / distinct X across these reports" → collect candidates from
  EVERY report read (not only those linked to the anchor entity), dedupe by
  canonical name, list the whole set.
- Timeline questions → collect date statements from the chunk texts;
  first-seen claims need the earliest stated date, not the most detailed report.
- Lineage questions → resolve each variant entity, read the chunk introducing
  it for the capability delta.
- Targeting/corroboration questions → group the chunks by vendor; a claim
  backed by multiple vendors is stronger than a single-source one.
\end{lstlisting}
\end{skillpromptbox}

\end{document}